\documentclass[12pt, leqno]{amsart}
\usepackage{amsmath}
\usepackage{amssymb}
\usepackage{algorithm}
\usepackage[noend]{algpseudocode}

\usepackage{bm}
\usepackage{graphicx}
\usepackage{caption}
\usepackage{subcaption}
\usepackage[hidelinks]{hyperref}
\usepackage{mathtools}
\usepackage{comment}
\usepackage{enumerate} 
\usepackage{parskip}
\usepackage{hyperref}

\usepackage{xcolor}
\allowdisplaybreaks

\newtheorem{theorem}{Theorem}[section]

\theoremstyle{definition}

\theoremstyle{remark}
\newtheorem{remark}[theorem]{Remark}
\numberwithin{equation}{section}

\newcommand{\R}{\mathbb{R}}

\newcommand{\E}{\mathbb{E}}

\numberwithin{equation}{section}

\begin{document}
\title[] 
{Neural solver for Wasserstein Geodesics and optimal transport dynamics}
\author{Yan-Han Chen}
\thanks{Department of Statistics, Iowa State University, Ames, IA 50011, USA}
\email{yanhanc@@iastate.edu}

\author{Hailiang Liu}
\thanks{Department of Mathematics, Iowa State University, Ames, IA 50011, USA}
\email{hliu@iastate.edu}



\date{\today}
\subjclass{93E20, 49Q22}
\keywords{Wasserstein geodesics, optimal transport, neural networks, data-driven methods}

\begin{abstract} In recent years, the machine learning community has increasingly embraced the optimal transport (OT) framework for modeling distributional relationships. In this work, we introduce a sample-based neural solver for computing the Wasserstein geodesic between a source and target distribution, along with the associated velocity field. Building on the dynamical formulation of the optimal transport (OT) problem, we recast the constrained optimization as a minimax problem, using deep neural networks to approximate the relevant functions. This approach not only provides the Wasserstein geodesic but also recovers the OT map,  enabling direct sampling from the target distribution. By estimating the OT map, we obtain velocity estimates along particle trajectories, which in turn allow us to learn the full velocity field.  The framework is flexible and readily extends to general cost functions, including the commonly used quadratic cost. We demonstrate the effectiveness of our method through experiments on both synthetic and real datasets.
\end{abstract}

\maketitle


\section{Introduction}
In recent years, the machine learning community has shown growing interest in the optimal transport (OT) framework \cite{villani2021topics}, which seeks to determine the most cost-effective transformation between probability distributions. 
Beyond yielding an optimal transport plan, OT also induces a natural metric on the space of probability measures, the Wasserstein distance, which  provides a geometrically meaningful way to compare distributions. This distance has proven useful in a wide range of  applications, including generative modeling \cite{arjovsky2017wasserstein, rout2021generative, fan2023neural, korotinwasserstein}, domain adaptation \cite{seguy2018large, courty2016optimal, courty2017joint, xie2019scalable},  and computational geometry \cite{peyre2019computational, S15}. 

When an optimal transport map exists, its computation can be formulated as a fluid dynamics problem that minimizes kinetic energy, as introduced by Benamou and Brenier \cite{benamou2000computational}. This dynamic formulation has deepened our understanding of Wasserstein geodesics, which describes the continuous evolution of distributions along optimal transport paths.
The richer information provided by this perspective not only helps the design of  efficient sampling methods from high dimensional distributions \cite{fin20}, but also makes learning dynamical Wasserstein geodesics practically valuable in real-world settings. 

In many applications -- especially in machine learning and data science-- only empirical samples from the underlying distributions are available, 
showing the need for sample-based methods (see, e.g., \cite{fan2023neural, liu2021learning,seguy2018large}).
However, accurately estimating Wasserstein geodesics and associated transport maps from finite samples remains a major challenge, especially in high-dimensional settings. Despite the recent progress in sampled-based optimal transport methods, several fundamental issues such as interpolation accuracy, computational efficiency, and scalability continue to be actively investigated. 
The main objective of this article is to develop a sample-based learning framework using a Lagrangian description of dynamical OT to compute the Wasserstein geodesic between source and target distributions, along with the associated velocity field.

Below we briefly review 
the relevant historical developments and motivate our proposed work.
\subsection{Optimal Transport}
The optimal transport (OT) problem, originally formulated
by Monge in 1781 \cite{monge1781memoire}, seeks the optimal strategy to relocate the mass from a
source distribution to a target distribution while minimizing the transportation cost. 
Specifically,
given two nonnegative measures $\mu_a$ and $\mu_b$
on $\R^d$ having
total equal mass (often assumed to be probability distributions), 
Monge's problem is defined as the following constrained optimization: 
\begin{subequations}\label{OMT}
    \begin{align}
   \min_{T}&\int_{\R^d}c(x,T(x))\mu_a(dx)\label{OMT_obj},\\
    \text{subject to}\quad  & T_\#\mu_a=\mu_b.\label{OMT_constr}
    \end{align}
\end{subequations}
Here, $T:\R^d\rightarrow
\R^d$ is a measurable transport map, and  $c(x,y)$ denotes the cost of moving a unit mass from point $x$ to point $y$, commonly taken to be of the form $c(x,y) = h(x-y)$ for a strongly convex function $h$. The constraint (\ref{OMT_constr}) guarantees that the pushforward of 
$\mu_a$ under $T$ matches the target 
distribution $\mu_b$. The pushforward measure  $T_\#\mu_a$ is defined 
such that for any measurable set $E\in \R^d$,  
$T_\#\mu_a(E) = \mu_a(T^{-1}(E))$, or equivalently, if $ X\sim\mu_a$, then  $T(X)\sim \mu_b$.
Monge's formulation is inherently challenging  
since an optimal transport map may not exist under general setups, especially when $\mu_a$ is not absolutely continuous with respect to the Lebesgue measure.  
Furthermore, the highly nonlinear dependence of the transportation cost on the transport map $T$ makes the constrained optimization problem difficult. 

These challenges led to further developments. In 1942, Kantorovich \cite{kantorovitch1958translocation} 
proposed a relaxed formulation where rather than seeking an explicit transport map $T$, a joint distribution 
$\pi$, called transport plan, on the product space $\R^d\times \R^d$ is considered as a broader and more tractable framework for optimal transport. Specifically, let $\Pi(\mu_a,\mu_b)$ denote the set of all joint distributions with marginals $\mu_a$ and $\mu_b$, respectively. The Kantorovich formulation of the optimal transport problem is then given by 
\begin{equation}\label{kan}
\inf_{\pi\in\Pi(\mu_a,\mu_b)}\int_{\R^d\times \R^d}c(x,y)\pi(dxdy).
\end{equation}
With appropriate assumptions on the cost function $c(\cdot, \cdot)$ and the distributions $\mu_a$ and $\mu_b$, Monge's formulation (\ref{OMT}) of OT problem admits a unique optimal transport map $T^*$ \cite{ brenier1991polar,chen2021optimal,villani2021topics,villani2008optimal}.
Moreover, the corresponding optimal  transport plan: 
\begin{equation*}
    \pi^* := \left(\text{Id}\times T^*\right)_\# \mu_a
\end{equation*}
solves the Kantorovich problem (\ref{kan}) \cite{gangbo1996geometry,villani2008optimal}.
This connection between Monge and Kantorovich formulations is fundamental in optimal transport theory and forms the basis for many modern computational approaches \cite{peyre2019computational}, while applications in high dimensional setup still remain challenging.
\\

\subsection{Density Control/Stochastic Control}
Monge and Kantorovich's formulations of the optimal transport problem are inherently static, in the sense that they focus on determining the optimal allocation of mass from a source distribution to a target distribution -- that is, they address the question of ``what goes where", but not ``how to get there". However, understanding the intermediate states during transportation is crucial for many applications in various domains, including reconstructing developmental trajectories in cell reprogramming  \cite{schiebinger2019optimal}, shape warping in computational geometry \cite{su2015optimal}, and controlling swarm robotics deployment  \cite{inoue2020optimal, krishnan2018distributed}.

This gap can be addressed by Benamou \& Brenier's dynamic formulation \cite{benamou2000computational} that captures the continuous evolution of mass over time. For the quadratic cost function 
 $c(x,y) = \lVert x-y\rVert^2$, the dynamic OT problem can be reformulated as the following variational problem: 
\begin{subequations}\label{dOMT_pde}
    \begin{align}
    \min_{v,\rho}\int_0^{S}&\int_{\R^d}
    L\left(v(x,t)\right)
    \rho(x,t)dxdt,\label{obj_pde}\\
     \text{subject to} \quad &\frac{\partial\rho}{\partial t} = 
    -\nabla\cdot\left(v\rho\right),\label{cont_eqn}\\
         &\rho(\cdot,0)= \rho_a(\cdot)\label{ini_distr}, \\
         &\rho(\cdot,S)= \rho_b(\cdot).
    \end{align}
\end{subequations}
Here, $L(v) = \lVert v \rVert ^2$ is the Lagrangian associated with kinetic energy, and $\rho_a, \rho_b$ denote the densities of source and terminal distributions $\mu_a$ and $\mu_b$, respectively. The time-dependent velocity field $v=v(x, t)$ in problem (\ref{dOMT_pde}) specifies how to steer the mass density $\rho(x, t)$ from $\rho_a$ to $\rho_b$ over the time interval $[0, 1]$. The constraint equation (\ref{cont_eqn}) is the continuity equation, which ensures mass conservation and is widely used in the context of fluid dynamics. This dynamic formulation can also be interpreted in a probabilistic framework:  given (\ref{cont_eqn}) and (\ref{ini_distr}),  $\rho$ describes the evolution of the 
distribution of a random process $X(t)$ governed by 
the ordinary differential equation 
\begin{equation}
   \begin{gathered}\label{rdn_state}
    \frac{dX(t)}{dt} = v(X(t),t),\\
    X(0)\sim \rho_a.
\end{gathered} 
\end{equation}
In this view, the OT problem becomes a special case of a deterministic control problem, where the goal is to steer the distribution of the state $X(t)$ from the source law $\rho_a$ to the target $\rho_b$ 
 by optimally choosing the control vector field $v(x, t)$. 
Throughout the paper, we assume without loss of generality that the terminal time is $S=1$, since any finite horizon can be rescaled to this case.
 
The correspondence between OT and its dynamic
formulation extends beyond the quadratic cost and holds for a broader class of cost functions.  Suppose the cost function  $c$ satisfies:  
\begin{gather*}
c\left(x(0),x(1)\right) = \min_{v(\cdot,t)} \int_0^1 L\Bigl(v\bigl(x(t),t\bigr)\Bigr) dt,\\
   \frac{dx(t)}{dt} = v\left(x(t),t\right)
\end{gather*}
for some Lagrangian $L(\cdot)\in C^1(\R^d)$.
A notable example is $L(v) = \lVert v \rVert ^p$ with $p\geq 1$, which corresponds
to the well-known  Wasserstein-$p$ distance  
\cite{villani2021topics,villani2008optimal}.
By (\ref{rdn_state}) and viewing the spatial integral in (\ref{obj_pde}) over $\R^d$ as an expectation
over the random process $X(t)$,
the dynamic OT problem  (\ref{dOMT_pde}) 
can be equivalently reformulated as a stochastic control problem: 
\begin{subequations}\label{dOMT}
\begin{gather}
    \min_{v\in \mathcal{V}}\int_0^1 \E 
    L\Bigl(v\bigl(X(t),t\bigl)\Bigl)dt,\label{obj_f}\\
         v(X(t),t) = \frac{dX(t)}{dt},\label{cstr_1}\\
         X(0)\sim \rho_a,\label{cstr_ini} \\
         X(1)\sim \rho_b\label{cstr_targ}.
\end{gather}
\end{subequations}
Here, $\mathcal{V}$ denotes the space of admissible control policies. In many 
data-driven scenarios, the explicit forms of $\rho_a$ and $\rho_b$ are unknown, and only 
empirical samples are available.
In such cases, a key objective is to generate new samples that are consistent with the target distribution $\rho_b$, 
given samples from both  $\rho_a$ and $\rho_b$.  

In this work, we adopt a trajectory-based  characterization of the OT problem and reformulate 
the stochastic optimal control problem  (\ref{dOMT}) as a saddle point optimization. The associated functions are parameterized using neural networks. Once the optimization converges, we move forward to the second phase,  where the optimal velocity field is recovered through a supervised learning problem. 

The main contributions of this work are as follows:
\begin{itemize}
    \item We propose a learning framework to estimate the optimal map, optimal velocity field, Wasserstein distance and geodesic between two distributions given only through samples. 
    We also prove the consistency of the corresponding saddle point formulation.
    \item The algorithm naturally extends to general Lagrangian functions $L(v,x,t)$, including those relevant to  Wasserstein-$p$ distances. Importantly, it avoids the need to enforce the 1-Lipschitz constraint.
    \item We validate the effectiveness of the proposed algorithm through extensive 
    numerical experiments on both synthetic and real datasets, in high and low-dimensional settings, and across both classical and generalized  Lagrangian formulations.
\end{itemize}

\subsection{Related work}
Conventional methods for numerically computing the Wasserstein
distance and geodesic typically rely on domain discretization, which is often mesh-dependent
\cite{benamou2010two,gangbo2019unnormalized,li2018parallel}.
Although effective in low-dimensional settings, these approaches become computationally intractable as dimensionality increases -- so called the curse of dimensionality that frequently arises in real-world applications.  A common strategy for handling high-dimensional settings is to apply entropic regularization to OT \cite{Benamou2015iterative, cuturi2013sinkhorn}, which allows the problem to be efficiently  solved using the Sinkhorn algorithm \cite{NIPS2017_491442df}. 
However, the Sinkhorn algorithm faces challenges when dealing with measures supported on continuous domains or with large number of samples \cite{genevay2016stochastic} -- scenarios frequently encountered in  machine learning applications. 

Neural networks, renowned for their capability to approximate continuous functions \cite{barron1993universal,cybenko1989approx,de2021approx}, have been widely used in continuous OT algorithms to estimate transport maps. This approach  enables to scale OT to high-dimensional problems that are otherwise intractable for traditional discrete methods.
A major application of OT in this context is generative modeling -- the task of learning models that can generate new data samples resembling a given dataset. One of the most influential frameworks in this area is the Generative Adversarial Network (GAN) \cite{goodfellow2014generative}, which formulates the learning process as a min-max game between two neural networks: a generator and a discriminator. A notable extension is the Wasserstein GAN (WGAN) \cite{arjovsky2017wasserstein},
which incorporates OT theory by replacing the original GAN loss with the Wasserstein-1 distance, resulting in improved training stability and sample quality. However, enforcing the necessary 1-Lipschitz constraint on the discriminator has proven to be a significant challenge \cite{gulrajani2017improved}, prompting the immediate  extensions and regularization strategies aimed at improving WGAN's  performance \cite{gulrajani2017improved,miyato2018spectral,wei2018improving, ijcai2019p305}. Concurrently, several works (e.g., \cite{cao2019multi, liu2019wasserstein, liu2018two}) have sought to address the training instabilities in WGAN  and to generalize the OT framework to more broader applications within generative modeling.

Beyond GAN-based approaches, neural networks have also been employed to learn OT maps or transportation plans from finite samples. Some methods directly parameterize  the OT map $T$ as defined in the Monge formulation  (\ref{OMT}) (e.g., \cite{fan2023neural,xie2019scalable,leygonie2019adversarial}), while others model the stochastic transport plan 
$\pi$ in the relaxed Kantorovich formulation  (\ref{kan}) (e.g. \cite{korotin2023neural,seguy2018large,lu2020large}), 
which is more general since a deterministic map $T$ may not exist in all cases. When an optimal map does exist, additional techniques such as regularization \cite{korotin2021neural, seguy2018large}, or cycle-consistency constraints \cite{lu2020large} are often necessary to guide learning from transportation plans toward a well-defined and possibly unique solution. For the special case of a quadratic cost, an  optimal transport map $T$ exists under the further  assumption that the source distribution is absolutely continuous \cite{brenier1991polar}. In this setting,  the map can be expressed as the gradient of a convex function, known as the Kantorovich potential. This insight has motivated approaches that learn these  potentials using input-convex neural networks (ICNNs) \cite{amos2017input}, 
leading to an active line of work in this direction (e.g., 
\cite{makkuva2020optimal,fan2021scalable,korotinwasserstein}). However, most of these methods are static, focusing on solving single-shot OT problems between two fixed distributions, without modeling the interpolation path or the dynamics of the transportation process. 

In contrast, our method builds upon the dynamic Benamou–Brenier (BB) formulation (\ref{dOMT_pde}), which recasts OT as a fluid-flow problem and can be naturally framed as a saddle point optimization problem.
The optimality conditions of the BB dynamic formulation, in both the primal and dual spaces, are characterized by a coupled PDE system or their weak formulations, as explored in \cite{liu2021learning} and \cite{pmlr-v244-gracyk24a}.
Instead of solving the coupled PDE system directly, \cite{liu2021learning} took advantage of the strict convexity of the Lagrangian $L(v)$ to propose a bidirectional learning framework, demonstrating scalability to high-dimensional settings.  Meanwhile \cite{pmlr-v244-gracyk24a} introduced GeONet, a mesh-invariant neural operator that learns nonlinear mapping from endpoint distributions to the entire geodesic under the classical quadratic cost by training against the coupled PDE system.  

In contrast to these approaches, our framework tackles the saddle point problem directly from a stochastic optimization perspective,  making it naturally compatible with general Lagrangians that may depend on the velocity field, time, and trajectory history.

\subsection{Organization}
The paper is organized as follows. Section 1 reviews related work in optimal transport and density control. In Sections 2 and 3, we present  the formulation of our approach and address theoretical issues, including consistency. 
In Section 4, we validate our method through  
numerical experiments on both synthetic and 
realistic datasets. Finally, Section 5 offers concluding discussions and future directions.

\section{Method} \label{phase1}
In this and next section, we present a data-driven framework for learning 
the optimal velocity field $v$ in equation (\ref{dOMT}), given samples $Z\sim\rho_a$ and $Y\sim\rho_b$. The approach involves two main phases.

In the first phase, we focus on finding the optimal transport map and determining the particle trajectory $X(t)$. This is formulated as a saddle-point problem for the original task. We approximate the transport map using a parameterized deep neural network.
Using this 
learned transport map, we derive an estimate of the velocity along the trajectory. However, this estimation
only provides partial information, as our goal is to recover the full velocity field. 

The second phase, described in Section \ref{phase2}, builds on this  trajectory-based velocity estimate
from the first phase to learn the complete velocity field. 

We begin by describing the first phase in detail.   

\subsection{Trajectory characterization} 
To satisfy the initial state constraint in (\ref{cstr_ini}), we reparameterize the trajectory $X(t)$ as 
\begin{equation*}
    X(t) = G(t;Z):=Z + tF(Z,t),~~ 0\le t\le 1, 
\end{equation*}
where $F: \R^d\times [0,1]  \rightarrow  \R^d$ is a function to be learned. This formulation ensures that the initial condition  
$$
X(0)= G(0;Z)=Z\sim\rho_a
$$
is satisfied by construction.  Here,  the function $G$ is fully and explicitly determined by $F$, which becomes the primary function of  learning.  

Along the trajectory $X(t) = G(t;Z)$, the corresponding  velocity field is given by the time derivative  
$$
v(X(t),t)=\frac{d G(t;Z)}{dt}.
$$
Based on this formulation, 
we define the following constrained
optimization problem,  which corresponds to  problem (\ref{dOMT}).
\begin{subequations}\label{dOMT_wk}
\begin{gather}
     \inf_{F} \int_0^1 \E 
    L\left(\frac{d G(t;Z)}{dt}\right)dt, \\ 
     \text{subject to } G(1;Z)\sim\rho_b,\label{cstr_targ_wk}\\
     G(t;Z)=Z + tF(Z,t).
\end{gather}
\end{subequations}
To enforce the constraint (\ref{cstr_targ_wk}), we evaluate  
the discrepancy between the target distribution $\rho_b$ and the distribution of the particles mapped by $G(1;Z)$. A common choice for the metric is the 
Wasserstein-1 distance,  $W(\cdot, \cdot)$, which quantities the minimal transport cost between two probability measures. 
Notably, $W\left(\rho_1,\rho_2\right)=0$ if and only if $\rho_1 = \rho_2$ almost everywhere.
By replacing the terminal constraint (\ref{cstr_targ_wk}) with the condition 
$$
W(\rho_b,G(1;\cdot)_{\#}\rho_a)=0,
$$ 
where $G(1;\cdot)_{\#}\rho_a$ denotes the pushforward of $\rho_a$ under $G(1;\cdot)$, we obtain  
an equivalent formulation of the optimization problem:
\begin{equation}
    \begin{gathered}
    \inf_{F} \int_0^1 \E 
    L\left(\frac{d G(t;Z)}{dt}\right)dt, \\ 
     \text{subject to }  W\left(\rho_b, G(1;\cdot)_{\#}\rho_a\right) =0,
     \\  G(t;Z)=Z + tF(Z,t).
\end{gathered}\label{Problem_W}
\end{equation}
A common strategy for solving such constrained optimization problems is 
to introduce a Lagrange multiplier
$\lambda$ and reformulate the problem as a saddle-point optimization: 
\begin{equation}
\begin{gathered}
    \sup_{\lambda} \inf_F \int_0^1 \E 
    L\left(\frac{d G(t;Z)}{dt}\right)dt+
    \lambda W\left(\rho_b,G(1;\cdot)_{\#}\rho_a\right),\\
    \text{subject to } G(t;Z)=Z + tF(Z,t).\\
\end{gathered}\label{Problem_minmax}
\end{equation}

\begin{remark}
In the Wasserstein geodesic formulation, $F(Z,t)$ can be chosen to be $t$-independent in cases where $L$ does not explicitly depend on $x$ and is solely a convex function of $v$, such as $L(v)=\frac{1}{2}\|v\|^2$. 
This choice is expected to simplify both the algorithm and learning procedure. In this work, we aim to also handle the general case with $L=L(v, x, t)$, for which  
$F=F(Z, t)$ is necessary.
The general case will be further discussed later in Section \ref{general_Lagr}.
\end{remark}
\subsection{Kantorovich-Rubinstein Duality}
Since we only have access to samples, directly computing the Wasserstein distance  $W(\cdot, \cdot)$ in the objective function (\ref{Problem_minmax}) is intractable.  
To overcome this, we leverage Kantorovich-Rubinstein Duality \cite{villani2008optimal}, which gives a sample-based formulation of the Wasserstein-1 distance. 
 
Let $\hat{Y}:=G(1;Z)$ denote a random sample from the pushforward 
distribution $G(1;\cdot)_{\#}\rho_a$.
Then,the Wasserstein-1 distance $W(\rho_b, G(1;\cdot)_{\#}\rho_a)$ can then be expressed as:   
\begin{equation}\label{KRduality}
W\left(\rho_b,G(1;\cdot)_{\#}\rho_a\right) =
    \sup_{\phi: \text{1-Lip}}  
\mathbb{E}\left[\phi(\hat{Y})\right] - \mathbb{E}\left[\phi(Y)\right]=
     \sup_{\phi: \text{1-Lip}}
\mathbb{E}\left[\phi\left(G(1;Z)\right)\right]-
   \mathbb{E}\left[\phi(Y)\right],
\end{equation}
where the supremum is taken over all 1-Lipschitz functions $\phi$.  

To further simplify the formulation, 
we introduce a random variable 
$U$ uniformly distributed in $[0,1]$, independent of $Z$ and $Y$. Incorporating this into the saddle-point problem, we rewrite the objective: 
\begin{align*}
     & \sup_{\lambda}\inf_{F} 
    \int_0^1 \E     
    L\left(\frac{d G(t;Z)}{dt}\right)dt
    +\lambda W\left(\rho_b,G(1;\cdot)_{\#}\rho_a\right)
\end{align*}
  as   
  \begin{align*}
     & \sup_{\lambda}\inf_{F} 
    \mathbb{E}_{Z\sim\rho_a,U\sim U(0,1)}\left[
     L\left(\frac{d G(U;Z)}{dt}\right)
    \right]+\lambda\sup_{\phi: \text{1-Lip}}
\left(\mathbb{E}\left[\phi\left(G(1;Z)\right)\right]-    \mathbb{E}\left[\phi(Y)\right]\right).
\end{align*} 
Assuming the order of optimization can be exchanged, we arrive at the 
following saddle-point formulation: 
\begin{align}\label{lip_merge}    
    & \inf_{F}\sup_{\lambda}\sup_{\phi: \text{1-Lip}}
    \mathbb{E}\left[
     L\left(\frac{d G(U;Z)}{dt}\right)\right]+\lambda\left(
\mathbb{E}\left[\phi\left(G(1;Z)\right)\right]-    \mathbb{E}\left[\phi(Y)\right]\right).
\end{align} 
Finally, by absorbing the scalar Lagrangian multiplier $\lambda$ into the 1-Lipschitz function $\phi$, we obtain a more streamlined formulation in which  $\phi$ is treated as a general Lipschitz function. This leads to the following min-max problem:
\begin{subequations}\label{learn_F}
\begin{gather}
     \inf_{F}\sup_{\phi: \text{Lip}}
    \mathbb{E}\left[
     L\left(\frac{d G(U;Z)}{dt}\right)\right]+\left(
     \mathbb{E}\left[\phi\left(G(1;Z)\right)\right]-
    \mathbb{E}\left[\phi(Y)\right]\right),\label{learn_F_1} \\ 
     \text{subject to } 
     G(t;Z)=Z + tF(Z,t).
\end{gather}
\end{subequations}
The following result guarantees the theoretical consistency of this formulation.
\begin{theorem} 
Suppose the min-max problem (\ref{learn_F}) admits a unique solution $(F^*,\phi^*)$. Then $F^*$ is also a solution to problem (\ref{Problem_W}),  which is equivalent to the original problem (\ref{dOMT_wk}).
\end{theorem} 
The proof is given in Appendix \ref{pf1}. 

In practical implementations, the expectations in (\ref{learn_F}) typically do not admit closed-form expressions. Nevertheless, by the law of large numbers, these expectations can be approximated to arbitrary accuracy using sample means, provided that enough independent samples are available. 
\subsection{Deep neural network approximations}
To numerically solve the optimization problem (\ref{learn_F}), we approximate the functions
$F$ and $\phi$ using 
deep neural networks.

A fully connected feedforward neural network can be mathematically expressed as a composition of linear transformations and nonlinear activation functions. 
Specifically, a neural network 
$f_{\theta}:\R^{N_1}\rightarrow \R^{N_L}$ of depth $L-2$ is defined as 
$$
f_{\theta}(\cdot) := h_{L-1} \circ \sigma_{L-2} \circ h_{L-2}
\circ \dotsm \circ \sigma_1 \circ h_1, 
$$
where each $h_j:\R^{N_{j}}\rightarrow R^{N_{j+1}}$, $j=1,2....,L-1$ is a linear transformation of the form 
$$
h_j(x)=W_jx+b_j,
$$
with weight matrix  
$W_j\in \R^{N_{j+1}\times N_{j}}$ and bias vectors  $b_j\in \R^{N_{j+1}}$. The function  $\sigma_j:\R\rightarrow\R$ is a nonlinear activation function applied component-wise to the output of $h_j$. 
The set of all trainable parameters in the network is  represented by $\theta\in\R^N$, which includes all weights and biases:  $W_1,b_1,\dotsc,W_{L-1},b_{L-1}$, with 
$$
N=\sum_{j=1}^{L-1} \left(N_j+1\right)N_{j+1}.
$$ 
Common activation functions include the hyperbolic
tangent (tanh),  sigmoid, rectified linear unit (ReLU) and Leaky ReLU (LReLU) \cite{apicella2021survey}. 
In our setting, we use two deep neural networks $F_\theta:\R^{d}\to\R^{d}$ and 
$\phi_\omega:\R^{d}\to\R$,  parameterized by
$\theta$ and $\omega$, respectively, to approximate $F$ and $\phi$. We select $L=5$ for both $F_\theta$ and $\phi_\omega$.
For the activation function $\sigma_j$ in $F_\theta$ and $\phi_\omega$, we choose Leaky ReLU function with negative slope $0.2$ throughout the experiments in Section \ref{experiment}. 
The formulation of this Leaky ReLU is
\begin{equation*}
    \text{LReLU}(x) = \begin{cases}
			x, & \text{if $x\ge0$}\\
            0.2 x, & \text{otherwise.}
		 \end{cases}
\end{equation*}

Given independent samples $\{Z_i\}_{i=1}^n\sim\rho_a$ and 
$\{Y_i\}_{i=1}^n\sim\rho_b$, we
discretize and approximate the objective in problem (\ref{learn_F}) using empirical averages. Specifically, with  $ \{U_j\}_{j=1}^m$ as independent samples drawn from $U(0,1)$, we propose the following finite-sample formulation 
\begin{equation}
    \begin{gathered}
       \min_{\theta}\max_{\omega}
    \frac{1}{nm}\sum_{j=1}^m\sum_{i=1}^n\left[
     L\left(\frac{d G_\theta(U_j;Z_i)}{dt}\right)\right]+\left(
     \frac{1}{n}\sum_{i=1}^n
     \left[\phi_\omega\left(G_\theta(1;Z_i)\right)\right]-
     \frac{1}{n}\sum_{i=1}^n\left[\phi_\omega(Y_i)\right]\right), \\ 
     \text{subject to } 
     G_\theta(t;Z_i)=Z_i + tF_\theta(Z_i,t).
\end{gathered}\label{learn_Ftheta}
\end{equation}
The proposed algorithm for solving this min-max problem is provided in Algorithm \ref{algorithm_F}.
\begin{remark}
In fact, implementing the algorithm does not require the sample sets drawn from the two distributions $\rho_a$ and $\rho_b$ to have the same size, nor does it require the samples to be paired. 
This is because the second term in \eqref{learn_Ftheta}, where the terminal constraint is enforced weakly, independently approximates the expectations with respect to each distribution. This observation also applies to Algorithm \ref{algorithm_F} presented later. 
\end{remark}

\subsection{Lipschitz constraint on the critic neural network}
Our learning problem (\ref{learn_Ftheta}) falls within the framework of Generative Adversarial Networks (GANs) \cite{goodfellow2014generative}, where the actor network ($F_\theta$) and the critic network ($\phi_\omega$) are adversarially trained. 
A key distinction in our setup is the application of a 
 Lipschitz constraint on the critic function in (\ref{learn_F_1}), derived from the use of the Wasserstein-$1$ distance. Importantly, our framework enforces a general Lipschitz constraint, rather than the stricter 1-Lipschitz condition commonly seen in Wasserstein GANs \cite{arjovsky2017wasserstein}.  Many subsequent works \cite{amos2017input,korotinwasserstein,fan2021scalable,makkuva2020optimal} have addressed the challenges posed by enforcing this non-trivial constraint. To implement the Lipschitz condition in our parameterized critic network, we follow the approach in \cite{miyato2018spectral}, applying  spectral normalization to each weight matrix $\tilde{W}_j$ of $\phi_\omega$, and selecting 1-Lipschitz activation functions $\sigma_j$. This ensures that the resulting network is 1-Lipchitz. We then scale the output by a learnable parameter, inspired by the derivation of our method between (\ref{lip_merge}) and (\ref{learn_F}), to generalize beyond the 1-Lipschitz case.   
 
 Finally, the critic network $\phi_\omega$ is defined as 
\begin{align*}
\phi_{\omega}(\cdot) :=\lambda \cdot \tilde{h}_{m-1} \circ \sigma_{m-2} \circ \tilde{h}_{m-2}
\circ \dotsm \circ \sigma_1 \circ \tilde{h}_1,    
\end{align*}
where each transformed layer $\tilde{h}_j(x)=(\tilde{W}_jx+b_j)/ \lVert \tilde{W}_j \rVert_2 $, and $ \lVert \tilde{W}_j \rVert_2$ is the spectral norm of $\tilde{W}_j$. Again, Leaky ReLU with negative slope $0.2$ is used as the activation throughout the experiments in Section \ref{experiment}.

\section{Velocity field recovery} \label{phase2}
\subsection{Velocity field recovery}
Keep in mind that in problem (\ref{dOMT}), a key objective is to recover the full velocity field
$v(x,t)$ -- meaning we seek to determine the value of the vector field at any location  $x\in \R^d$ and time $t\in[0,1]$. However, the learning framework proposed for problem (\ref{learn_F})
 does not directly yield an explicit representation of $v(x,t)$.

To illustrate this, let $F_{\theta^*}$ denote the numerical solution to problem (\ref{learn_F}). The estimated trajectory  $G_{\theta^*}(t;z)$ starting at $z\in \R^d$ is given by 
$$
G_{\theta^*}(t;z) := z + tF_{\theta^*}(z,t).
$$ 
Using this learned
trajectory $G_{\theta^*}$ and equation  (\ref{cstr_1}), 
we can estimate the velocity
along the trajectory as   
\begin{equation*}
    v_{\theta^*}\left(G_{\theta^*}(t;z),t\right):=\frac{dG_{\theta^*}(t;z)}{dt}.
\end{equation*}
However, this only provides the values of the velocity along trajectories, not the full velocity field $v(x, t)$ where $x$ and $t$ are independent. 

To recover the complete velocity field $v(x, t)$, where the spatial and temporal arguments are independent, we introduce another neural network $v_\eta$ to
approximate the true time-varying velocity field. We treat 
$v_{\theta^*}\left(G_{\theta^*}(U;Z),U\right)$ as samples from 
the ground truth and train $v_\eta$ using  supervised learning. 

By employing the squared error loss, we formulate the following optimization problem to learn the velocity field $v$ for problem 
(\ref{dOMT}): 
\begin{equation}\label{learn_v}
    \min_{\eta}\frac{1}{mn}\sum_{j=1}^m\sum_{i=1}^n
    \left \lVert v_\eta(G_{\theta^*}(U_j;Z_i),U_j)-
    \frac{d G_{\theta^*}(U_j;Z_i)}{dt} \right \rVert^2.
\end{equation}
As our Phase 2, this procedure is described in Algorithm \ref{algorithm_V}. 
\begin{remark}
    Regarding the samples $Z_i$ and $U_j$ used for training in Phase 2, we recommend drawing fresh samples rather than reusing those from Phase 1. This helps prevent Phase 2 from simply memorizing velocities along the specific trajectories  learned in Phase 1.  In the synthetic experiments presented in Section~\ref{experiment}, where the true velocity field $v^*$ is available, 
    we observe no significant difference in the error 
    between $v_\eta$ and $v^*$ when comparing fresh samples to reused Phase 1 samples. However, this may be due to the relative simplicity of the synthetic settings, which makes the velocity field easier to learn. Moreover, since the trajectories learned in Phase 1 are treated as  "ground truth" during supervised training in Phase 2, a zero Phase 2 loss does not guarantee convergence of $v_\eta$ to $v^*$ (see Appendix \ref{detail_numerical} for more details).
\end{remark}
In summary, the proposed framework addresses problem (\ref{dOMT}) in two phases: first by solving (\ref{learn_Ftheta}) to estimate the geodesic, and then by solving (\ref{learn_v}) to learn the full velocity field.
It is worth noting that in some applications, the objective is to generate additional samples from $\rho_b$,  interpolate intermediate trajectories, or estimate velocities along the transport paths. 
In such cases, implementing Phase 1 to obtain the transport map is already sufficient, and Phase 2 -- aimed at recovering the Euler velocity field -- is not necessary.
\subsection{Extension to general Lagrangian}\label{general_Lagr}
Our method extends naturally to time-dependent optimal transport problems, in which the transportation cost also depends on the transport trajectory, rather than just the initial and terminal points. In such settings, a typical Lagrangian takes the form
\begin{equation*}
    L(x,v,t) = \frac{\lVert v\rVert^2}{2}-V(x),
\end{equation*}
where $V:\mathbb{R}^d\rightarrow\mathbb{R}$ is a potential function \cite{villani2008optimal}. 
For a general Lagrangian, the existence and uniqueness of an optimal solution are guaranteed when  $L(x,v,t)$ is strictly convex and superlinear in the velocity variable $v$ \cite{chen2016optimal,villani2008optimal,Figalli2008OptimalTransport, bernard2007optimal}.
For further details about the time-dependent optimal transport, we refer the reader to Villani \cite{villani2008optimal} and Chen et al. \cite{chen2016optimal}.
The Lagrangian in our previous setup (\ref{dOMT}) is a special case of this more general formulation: by selecting $V(x) = 0$, we recover the simpler form 
$ L(x,v,t) = L(v) =\frac{\lVert v \rVert^2 }{2}$. 
Using a similar approach to that in Section \ref{phase1}, we can numerically solve the more general version of  the dynamical optimal transport problem: 
\begin{subequations}\label{dOMT_generalL}
\begin{gather*}
    \min_{v}\int_0^1 \E 
    L\Bigl(X(t), v\bigl(X(t),t\bigl),t\Bigl)dt,\\
         v(X(t),t) = \frac{dX(t)}{dt},\\
         X(0)\sim \rho_a, \\
         X(1)\sim \rho_b. 
\end{gather*}
\end{subequations}
In practice, we implement this by replacing the simple kinetic energy with the more general Lagrangian:
\begin{equation*}
L\left( G_\theta(U_j;Z_i),\frac{d G_\theta(U_j;Z_i)}{dt},U_j\right)    
\end{equation*}
for the corresponding term in (\ref{learn_Ftheta}) and Algorithm \ref{algorithm_F}. A numerical experiment in this general setting will be given in Section \ref{experiment}.
It is worth noting that many existing methods based on the Benamou–Brenier dynamical formulation (e.g., \cite{liu2021learning,pmlr-v244-gracyk24a}) primarily focus on the setting where $L = L(v)$ depends solely on $v$ and is strictly convex. In such case,  the associated Hamiltonian $H(p):=\sup_v v\cdot p - L(v)$ admits an explicit form. 
In contrast, our framework naturally extends  to the more general class of Lagrangian considered above.
\begin{algorithm}
\caption{Training Procedure for Solving Problem (\ref{learn_Ftheta}).}
\label{algorithm_F}
\begin{algorithmic}[1]
\State \textbf{Input:} Sample size $n$; number of time samples $m$; Lagrangian $L(v)$.
\State Initialize networks $F_{\theta}$ and $\phi_{\omega}$. Define $G_\theta(t;z):= z+tF_{\theta}(z,t)$ for $(z,t)\in \R^d\times[0,1]$. 
\While{$\theta$ has not converged}
\State Sample $\{Z_i\}_{i=1}^{n} \overset{\mathrm{i.i.d}}{\sim} \rho_a$ and
$\{Y_i\}_{i=1}^{n} \overset{\mathrm{i.i.d}}{\sim} \rho_b$.
        \State Compute the objective function for
        $\phi_{\omega}$
        $$\ell_{\omega} \leftarrow \frac{1}{n}\sum_{i=1}^{n}\left[
        \phi_{\omega}\left(G_{\theta}(1;Z_i)\right)\right]
        -\frac{1}{n}\sum_{i=1}^{n}\left[
        \phi_{\omega}\left(Y_i\right)\right].$$
    \State Update $\omega$ by ascending $\ell_{\omega}$.
    \State Sample $\{Z_i\}_{i=1}^{n} \overset{\mathrm{i.i.d}}{\sim} \rho_a$ and
    $\{U_j\}_{j=1}^{m} \overset{\mathrm{i.i.d}}
    {\sim} U(0,1)$. 
    \State Compute the objective function for $F_\theta$
        $$
        \ell_{\theta} \leftarrow 
        \frac{1}{mn}\sum_{j=1}^m\sum_{i=1}^n
        L\left(  \frac{d G_{\theta}(U_j;Z_i)}{dt}\right)+
        \frac{1}{n}\sum_{i=1}^{n}\phi_{\omega}\left(G_{\theta}(1;Z_i)\right).
        $$
\State Update $\theta$ by descending $\ell_{\theta}$.
\State {\bf end while} 
\EndWhile    
\State \textbf{Output:} Trained geodesic $G_\theta$ and critic $\phi_\omega$.
\end{algorithmic}
\end{algorithm}

\begin{algorithm}
\caption{Training Procedure for Solving Problem (\ref{learn_v}).} 
\label{algorithm_V}
\begin{algorithmic}[1] 
\State \textbf{Input:} Trained neural network $G_{\theta^*}$ from Algorithm \ref{algorithm_F}.
\State Initialize the network $v_{\eta}$.
\While{$\eta$ has not converged}
 \State Sample $\{Z_i\}_{i=1}^{n} \overset{\mathrm{i.i.d}}{\sim} \rho_a$ and
    $\{U_j\}_{j=1}^{m} \overset{\mathrm{i.i.d}}
    {\sim} U(0,1)$.
\State Compute the object function  for $v_\eta$
\begin{equation*}
    \ell_{\eta} = \frac{1}{mn}\sum_{j=1}^m\sum_{i=1}^n\left\lVert
    v_{\eta}(G_{\theta^*}(U_j;Z_i),U_j)-
    \frac{d G_{\theta^*}(U_j;Z_i)}{dt}
    \right\rVert_2^2.
\end{equation*}
\State Update $v_{\eta}$ by descending $\ell_\eta$.
\State {\bf end while} 
\EndWhile
\State \textbf{Output:} Trained velocity field $v_\eta$.
\end{algorithmic}
\end{algorithm}
\section{Numerical Experiments}\label{experiment}
In this section, we evaluate our algorithms on both synthetic and
real datasets. 
The neural networks used for the transport map $G_\theta$, the critic $\phi_\omega$, and the velocity field $v_\eta$ each consists of three hidden layers and are trained using  RMSprop optimizer \cite{tieleman2012lecture}.
We adopt the Leaky ReLU activation function with a negative slope of $0.2$ across all experiments. 

Except for the general Lagrangian case in Synthetic-4, we use the standard quadratic Lagrangian $L(v)=\lVert v\rVert^2$. Under this choice,  the first term in the objective function (\ref{learn_Ftheta}) serves as an estimate of the squared Wasserstein-2 distance between the source and 
target distributions.

For the random variable $\{U_j\}_{j=1}^{m}$ used in Algorithm \ref{algorithm_F} and \ref{algorithm_V}, we adopt $m=10$ equally spaced time steps from $t = 0.1$ to $t = 1.0$ in Synthetic-1 through Synthetic-3. We set 
$m=20$ in Synthetic-4 and in the Real Data case.
In Synthetic-1 and Synthetic-3 where the analytical velocity field $v^*$ are available, we compare $v^*$ with the learned $v_\eta$ in the Appendix \ref{detail_numerical}.
For the real data scenario, we apply our algorithm to the MNIST handwritten digits dataset \cite{lecun1998gradient}, which serves as a high-dimensional benchmark. 
The velocity fields reported in this section are trained with fresh samples $Z_i$ and $U_j$. That is, new samples are drawn for the supervised training in Phase 2.

\noindent{\bf Synthetic Experiments} \\ 

\begin{figure}
    \centering
\vspace{-4pt}
\includegraphics[width=0.7\textwidth]{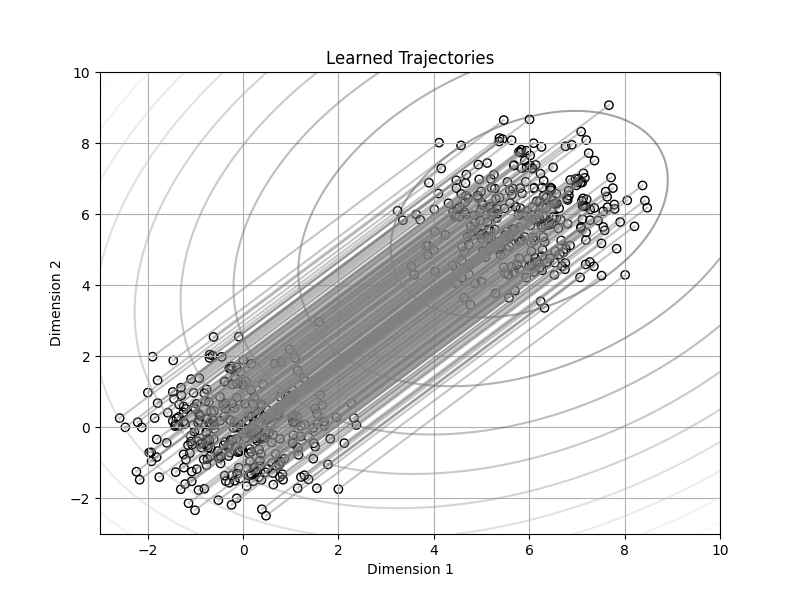}
    \captionsetup{aboveskip=-5pt} 
    \caption{Phase 1 of Synthetic-1: Samples of the source distribution $\rho_a$ (open circle at the bottom left corner) are
        transported to the upper right corner. 
        Each gray line illustrates the  learned trajectory $G_\theta(t;\cdot)_\#\rho_a$. 
        The contour shows the 
        true log density of $\rho_b$.}
    \label{fig:Gau2D_F}
\end{figure}

\begin{figure}
    \centering
\vspace{-4pt}
\includegraphics[width=0.8\linewidth]{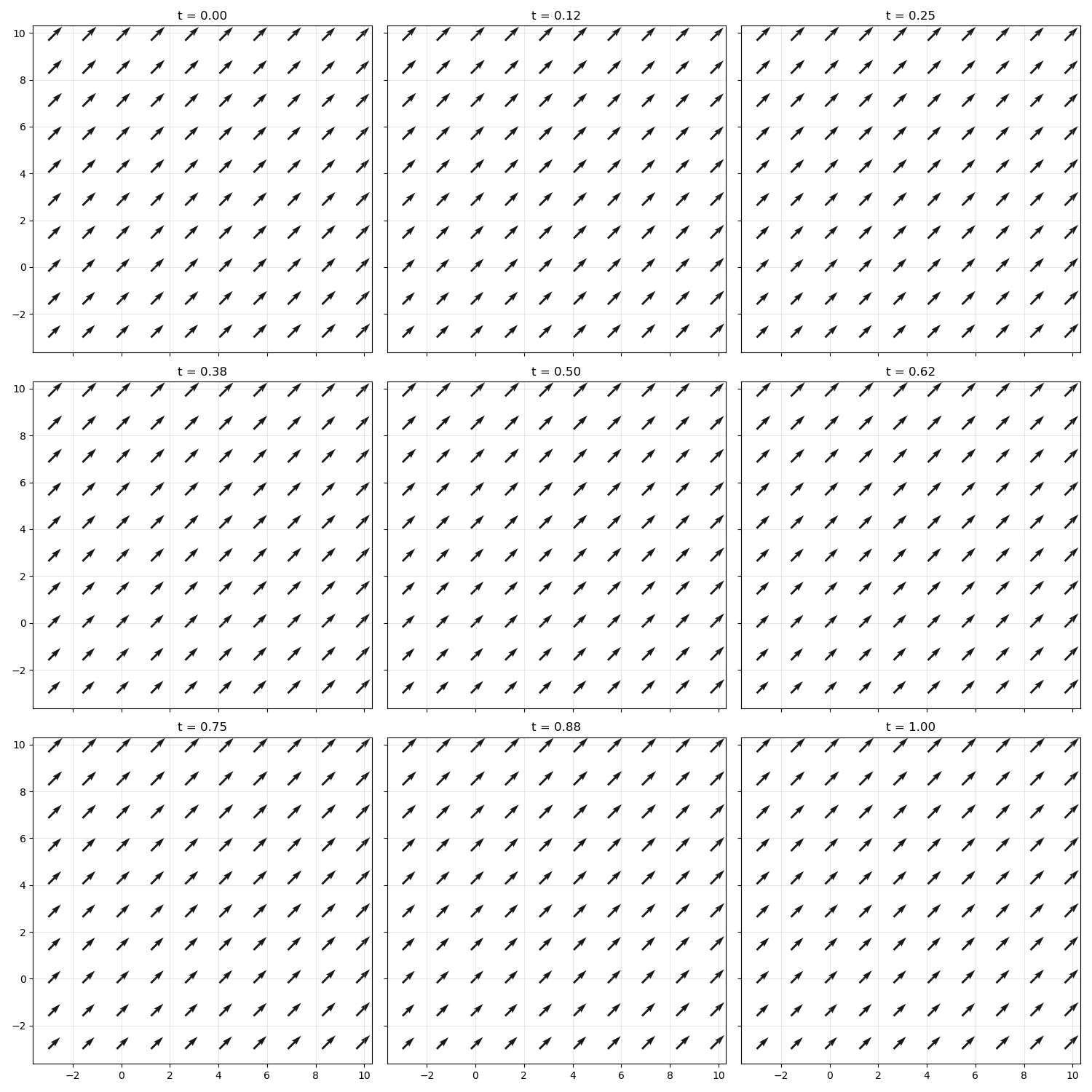}
    \caption{Phase 2 of Synthetic-1: learned velocity fields at selected time points. The arrows illustrate the direction and magnitude of the learned velocity.}
    \label{fig:Gau2D_v}
\end{figure}

\textbf{Synthetic-1: 2D-Gaussian.} 
In this 2D example, both $\rho_a$ and $\rho_b$ are Gaussian distributions
\begin{equation*}
   \rho_a= \mathcal{N}\left(\binom{0}{0},\begin{bmatrix}
1 & 0 \\
0 & 1 
\end{bmatrix}\right), \,\,\,\,
\rho_b= \mathcal{N}\left(\binom{6}{6},\begin{bmatrix}
1.5 & 0.5 \\
0.5 & 1.5 
\end{bmatrix}\right).
\end{equation*}
Figure \ref{fig:Gau2D_F} shows the learned trajectories of samples of $\rho_a$ in Phase 1. 
Figure \ref{fig:Gau2D_v} presents the learned velocity field at selected time points in Phase 2.

\begin{figure}
    \centering
\vspace{-4pt}
\includegraphics[width=0.7\textwidth]{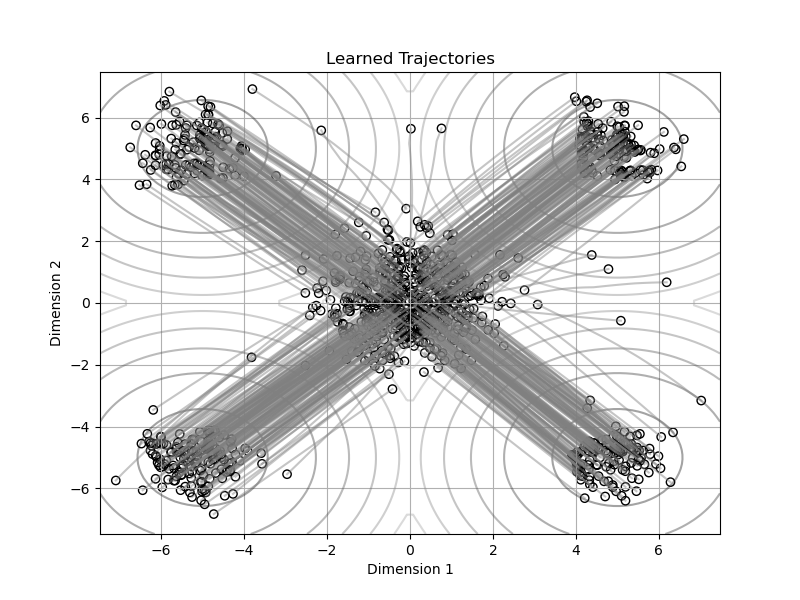}
    \captionsetup{aboveskip=-5pt} 
    \caption{Phase 1 of Synthetic-2: Samples of the source distribution $\rho_a$ (open circle at the center) are
        transported to the four corners. 
        Each gray line illustrates the  learned trajectory $G_\theta(t;\cdot)_\#\rho_a$. 
        The contour shows the 
        true log density of $\rho_b$.}
    \label{fig:MixG2D_F}
\end{figure}

\begin{figure}
    \centering
    \vspace{-4pt}
    \includegraphics[width=0.8\linewidth]{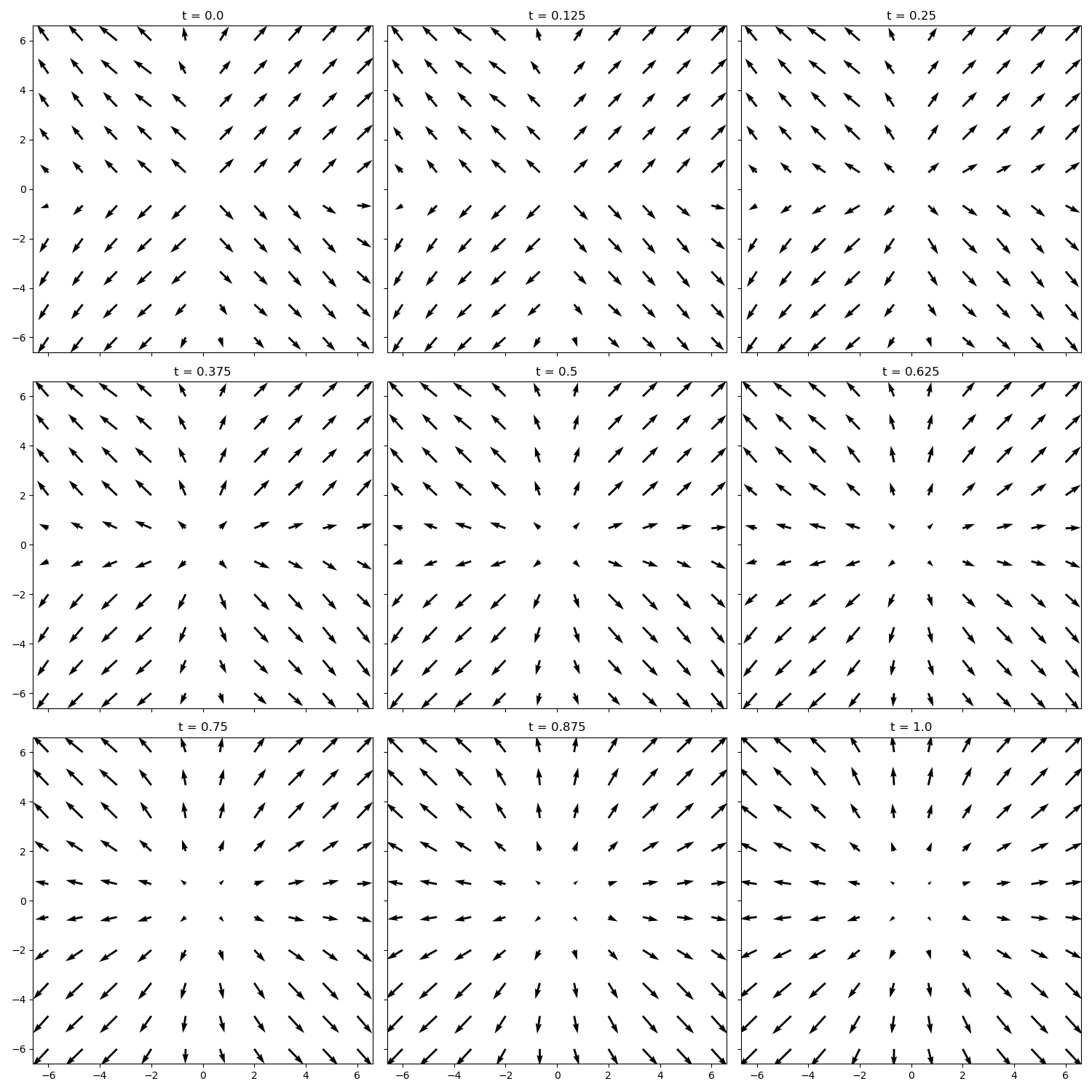}
    \caption{Phase 2 of Synthetic-2: learned velocity fields at selected time points. The arrows illustrate the direction and magnitude of the learned velocity.}
    \label{fig:MixG2D_v}
\end{figure}

\textbf{Synthetic-2: Mixture of 2D-Gaussians.} Here, $\rho_a$ is a standard 2-dimensional Gaussian, and 
$\rho_b$ is a mixture of 4 Gaussian components with the same covariance matrix as $\rho_a$, positioned at the corners of a square. Figure \ref{fig:MixG2D_F} and  \ref{fig:MixG2D_v} present the learned geodesic and velocity field. 

\begin{figure}
    \centering
\vspace{-4pt}
\includegraphics[width=0.7\textwidth]{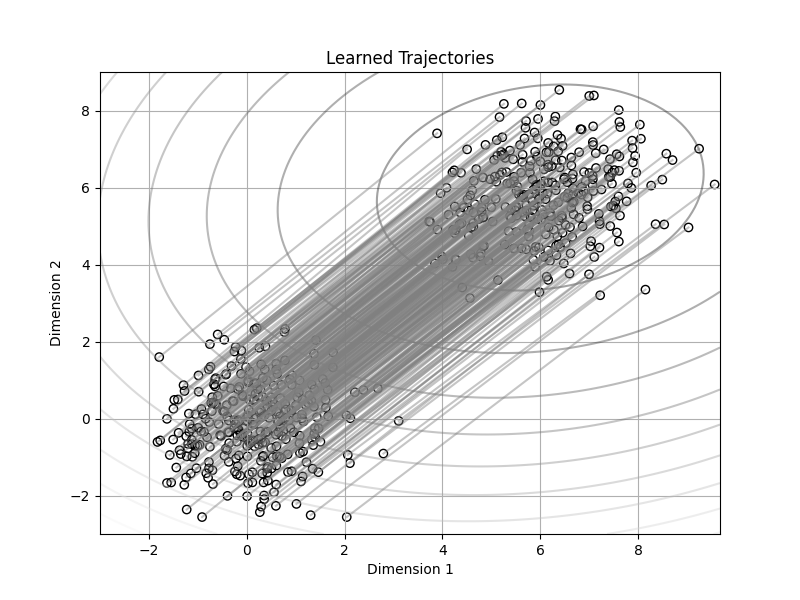}
    \captionsetup{aboveskip=-5pt} 
    \caption{Phase 1 of Synthetic-3:  2-dimensional projection of the 10-dimensional distributions. Samples of the source distribution $\rho_a$ (open circle at the bottom left corner) are
        transported to the upper right corner. 
        Each gray line illustrates the  learned trajectory $G_\theta(t;\cdot)_\#\rho_a$. 
        The contour shows the 
        true log density of $\rho_b$.}
    \label{fig:Gau10D_F}
\end{figure}

\begin{figure}
    \centering
    \vspace{-4pt}
    \includegraphics[width=0.8\linewidth]{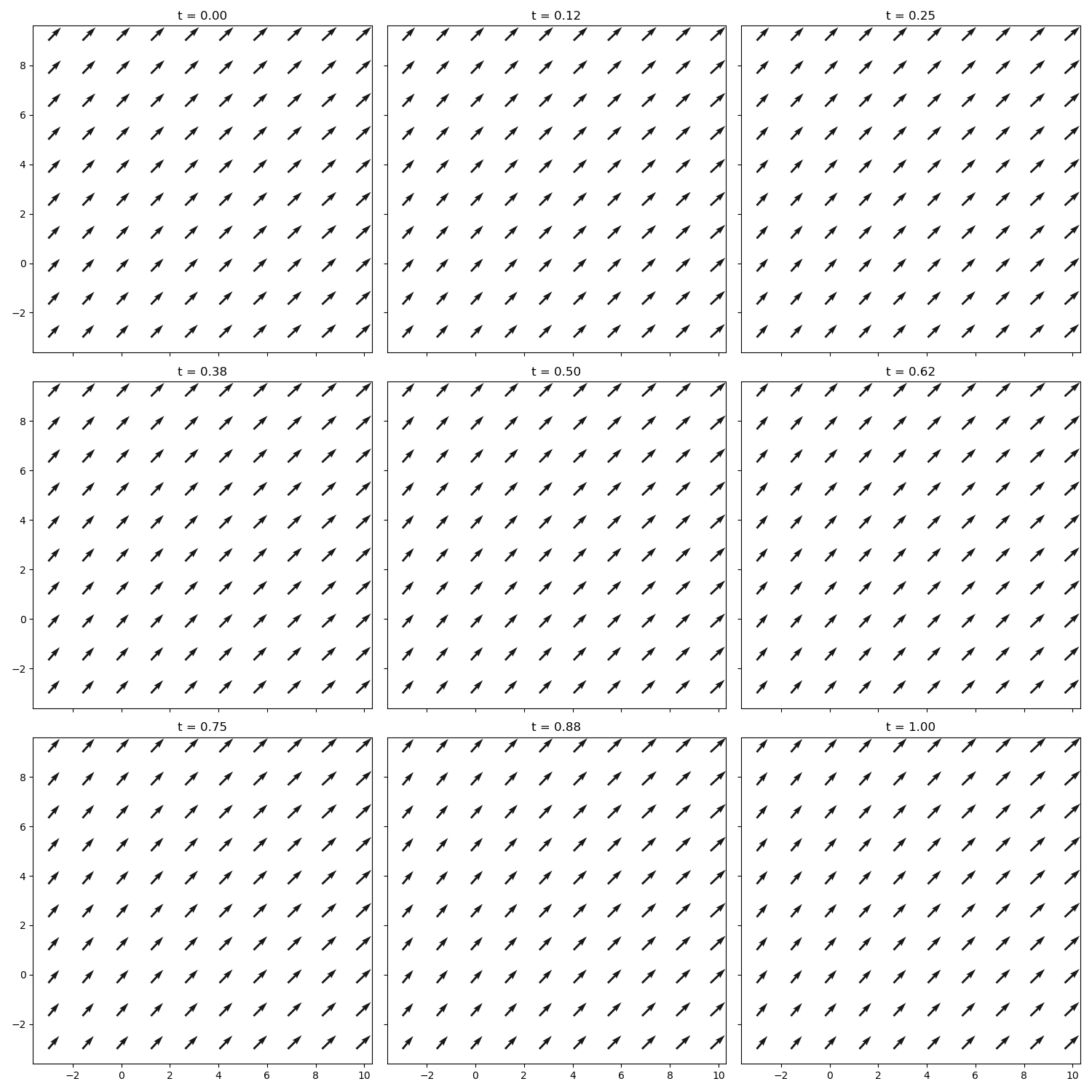}
    \caption{Phase 2 of Synthetic-3: learned velocity fields at selected time points. The arrows illustrate the direction and magnitude of the learned velocity.}
    \label{fig:Gau10D_v}
\end{figure}

\textbf{Synthetic-3: 10D-Gaussians.} 
For this experiment in the 10-dimensional space,  $\rho_a$ is chosen to be a standard Gaussian and $\rho_b$ another Gaussian. The learned geodesic and velocity field projected to the 2-dimensional space are shown in 
Figure \ref{fig:Gau10D_F} and \ref{fig:Gau10D_v}, respectively. 

\begin{figure}
     \centering
     \begin{subfigure}[b]{0.45\textwidth}
         \centering
         \includegraphics[width=\textwidth]{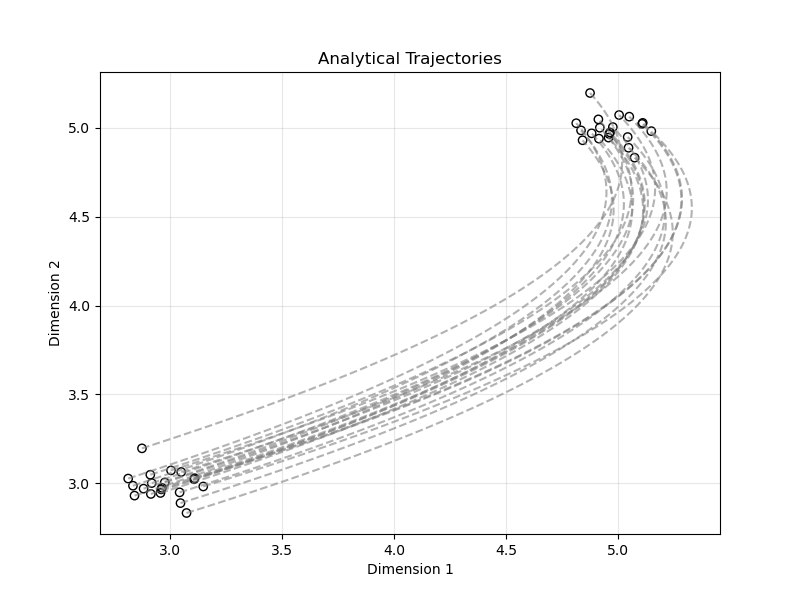}
     \end{subfigure}
     \begin{subfigure}[b]{0.45\textwidth}
         \centering
         \includegraphics[width=\textwidth]{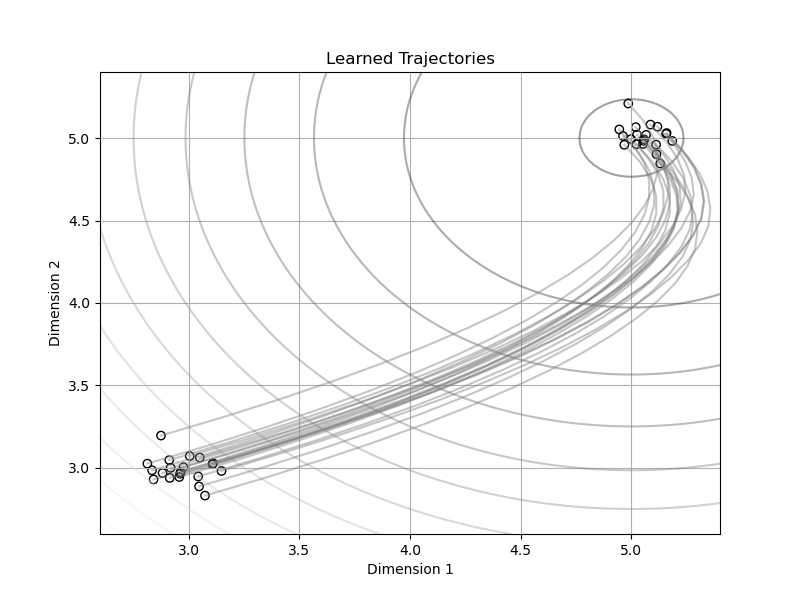}
     \end{subfigure}
        \caption{
        Phase 1 of Synthetic-4: 
        Left panel presents the analytical trajectories (\ref{anay_syn4_xt}) of 20 samples from $\rho_a$ (open circle at the bottom left corner). 
        In the right panel, those 20 samples are transported by the learned map,
        where the gray lines illustrate the learned trajectories $G_\theta(t;\cdot)_{\#}{\rho_a}$.
        The contour in the right panel shows the 
        true log density of $\rho_b$.
        } \label{fig:HO2D_F}
\end{figure}

\begin{figure}
    \centering
    \vspace{-4pt}
    \includegraphics[width=0.8\linewidth]{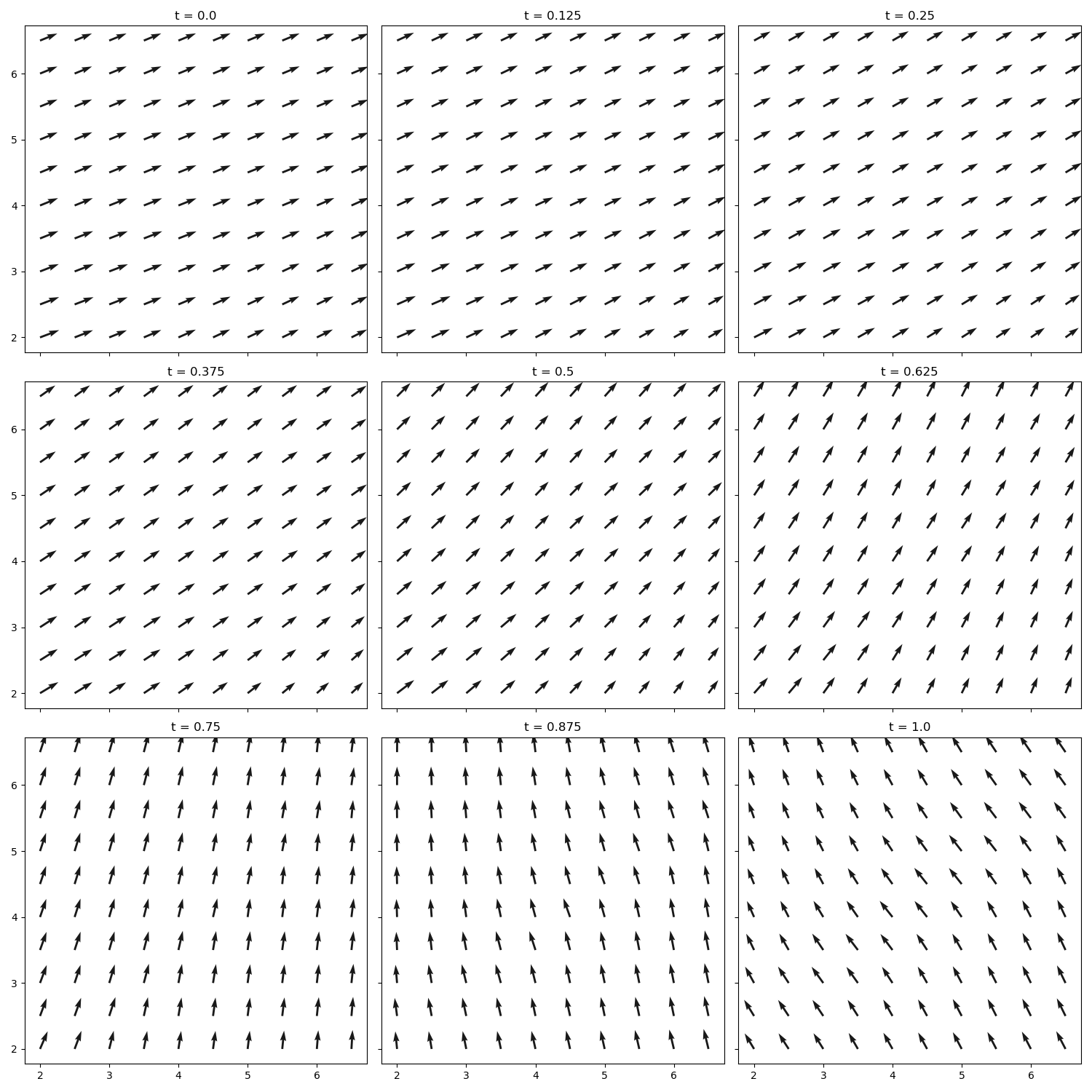}
    \caption{Phase 2 of Synthetic-4: learned velocity fields at selected time points. The arrows illustrate the direction and magnitude of the learned velocity.}
    \label{fig:HO2D_v}
\end{figure}

\textbf{Synthetic-4: 2D Harmonic Oscillation with general Lagrangian.}
   In this example, we consider the 2-dimensional harmonic oscillation. Specifically, for $x = (x_1, x_2)$ and $v$ in $\R^2$, 
   consider the Lagrangian 
   $$
   L(x,v) = \frac{1}{2} \left(\lVert v \rVert^2-\omega_1^2 x_1^2-\omega_2^2 x_2^2 \right),
   $$ 
   where $\omega_1, \omega_2>0$ are given. Solve the optimization problem:
\begin{gather*}
    \min_{v}\frac{1}{2}\int_0^1 \E \left[
   \lVert v\left(X(t),t \right) \rVert^2-\omega_1^2 X_1^2(t)-\omega_2^2 X_2^2(t)\right],\\
         v\left(X(t),t\right) = \frac{dX(t)}{dt},\\
         X(0)\sim \rho_a, \,\,
         X\left(1\right)\sim \rho_b.
\end{gather*}
We consider the case $\omega_1 = 1.2, \omega_2 = 0.1$ and the source and target distributions are defined as 
\begin{equation*}
   \rho_a= \mathcal{N}\left(\binom{m_{a,1}}{m_{a,2}},\begin{bmatrix}
0.01 & 0 \\
0 & 0.01 
\end{bmatrix}\right), \,\,\,\,
\rho_b= \mathcal{N}\left(\binom{m_{b,1}}{m_{b,1}},\begin{bmatrix}
0.01 & 0 \\
0 & 0.01 
\end{bmatrix}\right),
\end{equation*}
where $m_{a,1} = m_{a,2} = 3, m_{b,1} = m_{b,2} = 5$. In this setting, the analytical solution $X(t)$ is available
\begin{equation}
    \begin{aligned}\label{anay_syn4_xt}
     &X_1(t;x) =  \frac{\sin{(\omega_1(1-t))}+
     \sin{(\omega_1t)}}{\sin{\omega_1}}x_1+ \frac{\sin{(\omega_1t)}}{\sin{\omega_1}}(m_{b,1} - m_{a,1})\\
    &X_2(t;x) =  \frac{\sin{(\omega_2(1-t))}+
     \sin{(\omega_2t)}}{\sin{\omega_2}}x_2+ \frac{\sin{(\omega_2t)}}{\sin{\omega_2}}(m_{b,2} - m_{a,2})
\end{aligned}
\end{equation}
This analytical trajectory and the learned geodesic in Phase 1 are shown in Figure \ref{fig:HO2D_F}. In Figure \ref{fig:HO2D_v}, the velocity field learned in Phase 2 is shown.

\begin{figure}
     \centering
     \begin{subfigure}[b]{0.4\textwidth}
         \centering
         \includegraphics[width=\textwidth]{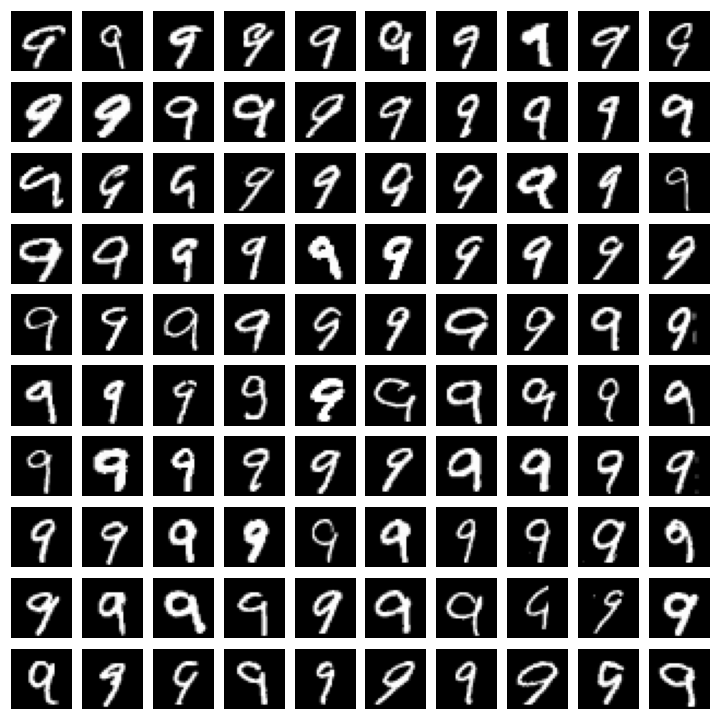}
     \end{subfigure}
     \begin{subfigure}[b]{0.4\textwidth}
         \centering
         \includegraphics[width=\textwidth]{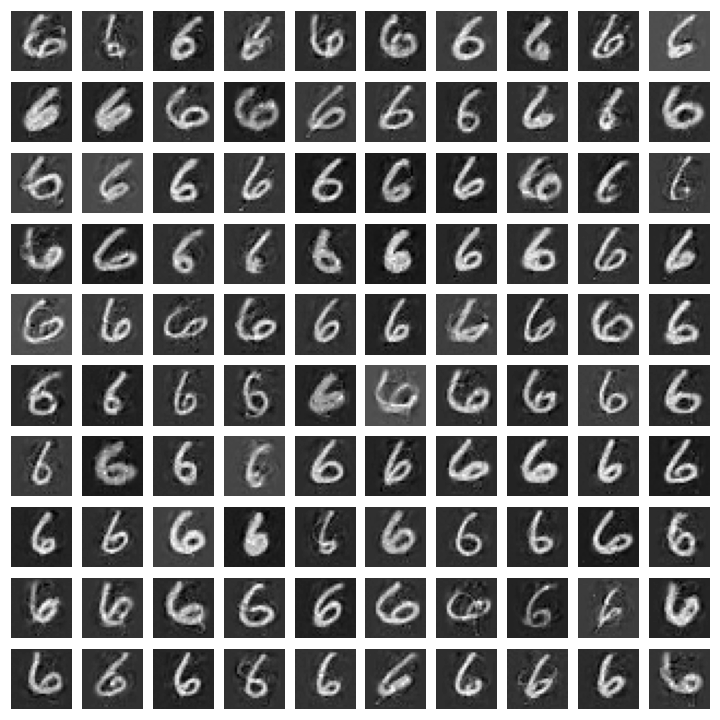}
     \end{subfigure}
        \caption{Phase 1 of  Real data (MNIST):
        Left panel contains 100 samples from the source distribution $\rho_a$ (digit 9). Right panel consists of transported samples $G_\theta(1;\cdot)_\#\rho_a$, which is meant to match $\rho_b$ (digit 6).
        }\label{fig:MNIST_F}
\end{figure}

\begin{figure}
    \centering
    \vspace{-5pt}
    \includegraphics[width=0.9\linewidth]{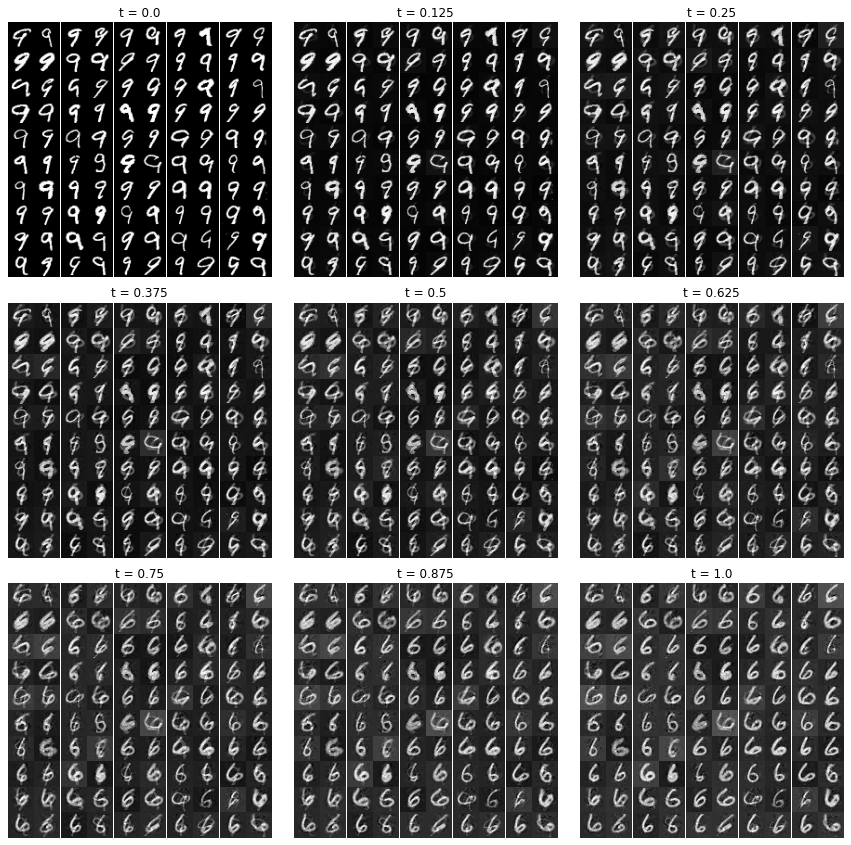}
    \captionsetup{aboveskip=0pt}
    \caption{Phase 1 of Real data case (MNIST): 
    Selected time slices of the transformation process from $\rho_a$ (digit 9) to $G_\theta(1;\cdot)_\#\rho_a$, which is meant to match $\rho_b$ (digit 6).}
   \label{fig:MNIST_Ft}
\end{figure}

\textbf{Real Data: MNIST digit transformation.} As a high-dimensional example ($28 \times 28$ dimensions), we consider the MNIST dataset. Specifically,  we aim to learn a transport map from digit 6 (representing $\rho_a$) to digit 9 (representing $\rho_b$).
Figure \ref{fig:MNIST_F} and  \ref{fig:MNIST_Ft} illustrate the learned geodesic trajectory in Phase 1.
\\

\textbf{Comment on the results of the numerical experiments.} 
We observed that the generated samples $G(1;\cdot)_{\#}\rho_a$ exhibit patterns similar to the target distribution $\rho_b$, as evidenced by comparing the empirical distribution of the pushed open circle to the true log density contour in synthetic cases, and by the convergence of the Wasserstein-1 distance toward zero during Phase 1 training in most cases (see Appendix \ref{detail_numerical} for details during training). Furthermore, visualization of  particle trajectories suggest that the transport of samples occurs efficiently. However, we note that the objective function values often  fluctuate during  Phase 1 training,  which may be due to the inherent instability of optimizing a min-max problem involving two adversarial networks. In some cases (e.g., Figure \ref{fig:MixG2D_F} of Synthetic-2), the mapped distribution does not fully match the target distribution, as indicated by the failure of the estimated Wasserstein-1 distance to converge to zero. 
We acknowledge the existence of additional regularization techniques for the critic function (e.g.,  \cite{gulrajani2017improved}) that might improve target distribution matching in Phase 1.
To address instability, alternative optimization methods may help improve performance. In addition, we observe that neural network  hyperparameters -- such as the choice of activation function, depth, width, and learning rate -- can significantly affect performance, including the quality of distribution matching. Thus,  additional regularization and careful hyperparameter tuning are two practical strategies for improving performance in cases where matching the target distribution and the pushfoward distribution is particularly challenging.
In Phase 2,  where the complete velocity field is learned, we observe that the discrepancy between the learned and the true velocity fields (e.g., Figure \ref{Gau10D_lossv} in Appendix \ref{detail_numerical}) does not fully vanish. This may stem from using the Phase 1 learned 
 trajectories as samples for ground truth, even though they are only approximations.
 
\section{Conclusion}
In this paper, we propose a sample-based learning framework to estimate the OT geodesic and corresponding velocity field for transporting the source distribution into the target distribution with minimum cost. 
Furthermore, the learning algorithm can be naturally applied to general Lagrangian where the transportation cost may depends on the velocity field, transport trajectory and time, including the classical case of simple kinetic energy.
We demonstrate the performance of our algorithm through a series of experiments including synthetic and realistic data, high-dimensional and low-dimensional setups, classical and general Lagrangian. 
\section*{Declarations}

\subsection*{Conflict of Interest}
The authors declare that they have no known competing financial interests or personal relationships that could have appeared to influence the work reported in this paper.

\subsection*{Data Availability}
The datasets generated during the the study are available from the corresponding author on reasonable request. The MNIST dataset used is a publicly available benchmark.
\medskip

\bibliography{ref}

@inproceedings{
korotin2023neural,
title={Neural Optimal Transport},
author={Alexander Korotin and Daniil Selikhanovych and Evgeny Burnaev},
booktitle={The Eleventh International Conference on Learning Representations },
year={2023}
}

@inproceedings{fin20,
  title     = {How to Train Your Neural ODE: The World of Jacobian and Kinetic Regularization},
  author    = {Finlay, Chris and Jacobsen, J{\"o}rn-Henrik and Nurbekyan, Levon and Oberman, Adam M},
  booktitle = {Proceedings of the 37th International Conference on Machine Learning (ICML)},
  year      = {2020}
}

@article{S15,
  author       = {Justin Solomon and Fernando de Goes and Gabriel Peyr{\'{e}} and Marco Cuturi
                  and Adrian Butscher and Andy Nguyen and Tao Du and Leonidas J. Guibas},
  title        = {Convolutional {W}asserstein Distances: Efficient Optimal Transportation on Geometric Domains},
  journal      = {ACM Transactions on Graphics},
  volume       = {34},
  number       = {4},
  pages        = {66:1--66:11},
  year         = {2015},
  publisher    = {ACM},
  doi          = {10.1145/2766963},
}

@article{monge1781memoire,
  title={M{\'e}moire sur la th{\'e}orie des d{\'e}blais et des remblais},
  author={Monge, Gaspard},
  journal={Mem. Math. Phys. Acad. Royale Sci.},
  pages={666--704},
  year={1781}
}

@article{benamou2000computational,
  title={A computational fluid mechanics solution to the {M}onge-{K}antorovich mass transfer problem},
  author={Benamou, Jean-David and Brenier, Yann},
  journal={Numerische Mathematik},
  volume={84},
  number={3},
  pages={375--393},
  year={2000},
  publisher={Springer-Verlag Berlin/Heidelberg}
}

@article{benamou2010two,
  title={Two Numerical Methods for the elliptic {M}onge-{A}mp{\`e}re equation},
  author={Benamou, Jean-David and Froese, Brittany D and Oberman, Adam M},
  journal={ESAIM: Mathematical Modelling and Numerical Analysis},
  volume={44},
  number={4},
  pages={737--758},
  year={2010},
  publisher={EDP Sciences}
}

@article{li2018parallel,
  title={A parallel method for earth mover’s distance},
  author={Li, Wuchen and Ryu, Ernest K and Osher, Stanley and Yin, Wotao and Gangbo, Wilfrid},
  journal={Journal of Scientific Computing},
  volume={75},
  number={1},
  pages={182--197},
  year={2018},
  publisher={Springer}
}

@article{gangbo2019unnormalized,
  title={Unnormalized optimal transport},
  author={Gangbo, Wilfrid and Li, Wuchen and Osher, Stanley and Puthawala, Michael},
  journal={Journal of Computational Physics},
  volume={399},
  pages={108940},
  year={2019},
  publisher={Elsevier}
}

@inproceedings{NIPS2017_491442df,
 author = {Altschuler, Jason and Niles-Weed, Jonathan and Rigollet, Philippe},
 booktitle = {Advances in Neural Information Processing Systems},
 pages = {},
 publisher = {Curran Associates, Inc.},
 title = {Near-linear time approximation algorithms for optimal transport via Sinkhorn iteration},
 url = {https://proceedings.neurips.cc/paper_files/paper/2017/file/491442df5f88c6aa018e86dac21d3606-Paper.pdf},
 volume = {30},
 year = {2017}
}

@inproceedings{arjovsky2017wasserstein,
  title={Wasserstein generative adversarial networks},
  author={Arjovsky, Martin and Chintala, Soumith and Bottou, L{\'e}on},
  booktitle={International Conference on Machine Learning},
  pages={214--223},
  year={2017},
  organization={PMLR}
}

@article{liu2021learning,
  title={Learning high dimensional {W}asserstein geodesics},
  author={Liu, Shu and Ma, Shaojun and Chen, Yongxin and Zha, Hongyuan and Zhou, Haomin},
  journal={arXiv preprint arXiv:2102.02992},
  year={2021}
}

@inproceedings{gulrajani2017improved,
  title={Improved training of {W}asserstein {GAN}s},
  author={Gulrajani, Ishaan and Ahmed, Faruk and Arjovsky, Martin and Dumoulin, Vincent and Courville, Aaron C},
  booktitle={Advances in Neural Information Processing Systems},
  pages = {},
  publisher = {Curran Associates, Inc.},
  volume={30},
  year={2017}
}

@article{cybenko1989approx,
  title={Approximation by superpositions of a sigmoidal function},
  author={Cybenko, George},
  journal={Mathematics of control, signals and systems},
  volume={2},
  number={4},
  pages={303--314},
  year={1989},
  publisher={Springer}
}

@article{barron1993universal,
  title={Universal approximation bounds for superpositions of a sigmoidal function},
  author={Barron, Andrew R},
  journal={IEEE Transactions on Information theory},
  volume={39},
  number={3},
  pages={930--945},
  year={1993},
  publisher={IEEE}
}

@article{de2021approx,
  title={On the approximation of functions by tanh neural networks},
  author={De Ryck, Tim and Lanthaler, Samuel and Mishra, Siddhartha},
  journal={Neural Networks},
  volume={143},
  pages={732--750},
  year={2021},
  publisher={Elsevier}
}

@article{brenier1991polar,
  title={Polar factorization and monotone rearrangement of vector-valued functions},
  author={Brenier, Yann},
  journal={Communications on Pure and Applied Mathematics},
  volume={44},
  number={4},
  pages={375--417},
  year={1991},
  publisher={Wiley Online Library}
}

@article{kantorovitch1958translocation,
  title={On the translocation of masses},
  author={Kantorovitch, Leonid},
  journal={Management Science},
  volume={5},
  number={1},
  pages={1--4},
  year={1958},
  publisher={INFORMS}
}

@article{gangbo1996geometry,
  title={The geometry of optimal transportation},
  author={Gangbo, Wilfrid and McCann, Robert J},
  journal={Acta Mathematica},
  volume={177},
  pages={113--161},
  year={1996},
  publisher={Springer}
}

@article{apicella2021survey,
  title={A survey on modern trainable activation functions},
  author={Apicella, Andrea and Donnarumma, Francesco and Isgr{\`o}, Francesco and Prevete, Roberto},
  journal={Neural Networks},
  volume={138},
  pages={14--32},
  year={2021},
  publisher={Elsevier}
}

@article{fan2023neural,
  title={Neural {M}onge map estimation and its applications},
  author={Fan, Jiaojiao and Liu, Shu and Ma, Shaojun and Zhou, Haomin and Chen, Yongxin },
  journal={Transactions on Machine Learning Research},
  year={2023},
  publisher={TMLR}
}

@inproceedings{krishnan2018distributed,
  title={Distributed optimal transport for the deployment of swarms},
  author={Krishnan, Vishaal and Mart{\'\i}nez, Sonia},
  booktitle={2018 IEEE Conference on Decision and Control (CDC)},
  pages={4583--4588},
  year={2018},
  organization={IEEE}
}

@inproceedings{makkuva2020optimal,
  title={Optimal transport mapping via input convex neural networks},
  author={Makkuva, Ashok and Taghvaei, Amirhossein and Oh, Sewoong and Lee, Jason},
  booktitle={International Conference on Machine Learning},
  pages={6672--6681},
  year={2020},
  organization={PMLR}
}

@inproceedings{amos2017input,
  title={Input convex neural networks},
  author={Amos, Brandon and Xu, Lei and Kolter, J Zico},
  booktitle={International Conference on Machine Learning},
  pages={146--155},
  year={2017},
  organization={PMLR}
}

@inproceedings{fan2021scalable,
  title={Scalable Computations of {W}asserstein Barycenter via Input Convex Neural Networks},
  author={Fan, Jiaojiao and Taghvaei, Amirhossein and Chen, Yongxin},
  booktitle={International Conference on Machine Learning},
  pages={1571--1581},
  year={2021},
  organization={PMLR}
}

@inproceedings{goodfellow2014generative,
    author = {Goodfellow, Ian J and Pouget-Abadie, Jean and Mirza, Mehdi and Xu, Bing and Warde-Farley, David and Ozair, Sherjil and Courville, Aaron and Bengio, Yoshua},
    title = {Generative adversarial nets},
    booktitle = {Advances in Neural Information Processing Systems},
    volume={27},
    pages = {},
    publisher = {Curran Associates, Inc.},
    year = 2014
}

@inproceedings{
miyato2018spectral,
title={Spectral Normalization for Generative Adversarial Networks},
author={Takeru Miyato and Toshiki Kataoka and Masanori Koyama and Yuichi Yoshida},
booktitle={International Conference on Learning Representations},
year={2018},
url={https://openreview.net/forum?id=B1QRgziT-},
}

@article{lecun1998gradient,
  title={Gradient-based learning applied to document recognition},
  author={LeCun, Yann and Bottou, L{\'e}on and Bengio, Yoshua and Haffner, Patrick},
  journal={Proceedings of the IEEE},
  volume={86},
  number={11},
  pages={2278--2324},
  year={1998},
  publisher={Ieee}
}

@article{tieleman2012lecture,
  title={Lecture 6.5-{RMSP}rop: Divide the gradient by a running average of its recent magnitude},
  author={Tieleman, Tijmen},
  journal={COURSERA: Neural Networks for Machine Learning},
  volume={4},
  number={2},
  pages={26},
  year={2012}
}

@article{chen2021optimal,
  title={Optimal transport in systems and control},
  author={Chen, Yongxin and Georgiou, Tryphon T and Pavon, Michele},
  journal={Annual Review of Control, Robotics, and Autonomous Systems},
  volume={4},
  number={1},
  pages={89--113},
  year={2021},
  publisher={Annual Reviews}
}

@article{chen2016optimal,
  title={Optimal transport over a linear dynamical system},
  author={Chen, Yongxin and Georgiou, Tryphon T and Pavon, Michele},
  journal={IEEE Transactions on Automatic Control},
  volume={62},
  number={5},
  pages={2137--2152},
  year={2016},
  publisher={IEEE}
}

@article{peyre2019computational,
  title={Computational optimal transport: With applications to data science},
  author={Peyr{\'e}, Gabriel and Cuturi, Marco},
  journal={Foundations and Trends in Machine Learning},
  volume={11},
  number={5-6},
  pages={355--607},
  year={2019},
  publisher={Now Publishers, Inc.}
}

@article{schiebinger2019optimal,
  title={Optimal-transport analysis of single-cell gene expression identifies developmental trajectories in reprogramming},
  author={Schiebinger, Geoffrey and Shu, Jian and Tabaka, Marcin and Cleary, Brian and Subramanian, Vidya and Solomon, Aryeh and Gould, Joshua and Liu, Siyan and Lin, Stacie and Berube, Peter and Lee, Lia and Chen, Jenny and Brumbaugh, Justin and Rigollet, Philippe and Hochedlinger, Konrad and Jaenisch, Rudolf and Regev, Aviv and Lander, Eric S.},
  journal={Cell},
  volume={176},
  number={4},
  pages={928--943},
  year={2019},
  publisher={Elsevier}
}

@article{inoue2020optimal,
  title={Optimal transport-based coverage control for swarm robot systems: Generalization of the {V}oronoi tessellation-based method},
  author={Inoue, Daisuke and Ito, Yuji and Yoshida, Hiroaki},
  journal={IEEE Control Systems Letters},
  volume={5},
  number={4},
  pages={1483--1488},
  year={2020},
  publisher={IEEE}
}

@article{su2015optimal,
  title={Optimal mass transport for shape matching and comparison},
  author={Su, Zhengyu and Wang, Yalin and Shi, Rui and Zeng, Wei and Sun, Jian and Luo, Feng and Gu, Xianfeng},
  journal={IEEE Transactions on Pattern Analysis and Machine Intelligence},
  volume={37},
  number={11},
  pages={2246--2259},
  year={2015},
  publisher={IEEE}
}

@article{cuturi2013sinkhorn,
  title={Sinkhorn distances: Lightspeed computation of optimal transport},
  author={Cuturi, Marco},
  journal={Advances in Neural Information Processing Systems},
  volume={26},
  year={2013},  
  pages = {2292–-2300}
}

@article{Benamou2015iterative,
author = {Benamou, Jean-David and Carlier, Guillaume and Cuturi, Marco and Nenna, Luca and Peyr\'{e}, Gabriel},
title = {Iterative {B}regman projections for regularized transportation problems},
journal = {SIAM Journal on Scientific Computing},
volume = {37},
number = {2},
pages = {A1111-A1138},
year = {2015},
}

@InProceedings{pmlr-v244-gracyk24a,
  title = 	 {GeONet: a neural operator for learning the {W}asserstein geodesic},
  author =       {Gracyk, Andrew and Chen, Xiaohui},
  booktitle = 	 {Proceedings of the Fortieth Conference on Uncertainty in Artificial Intelligence},
  pages = 	 {1453--1478},
  year = 	 {2024},
  volume = 	 {244},
  publisher =    {PMLR},
}

@inproceedings{seguy2018large,
  title={Large-Scale Optimal Transport and Mapping Estimation},
  author={Seguy, Vivien and Damodaran, Bharath Bhushan and Flamary, Remi and Courty, Nicolas and Rolet, Antoine and Blondel, Mathieu},
  booktitle={ICLR 2018-International Conference on Learning Representations},
  pages={1--15},
  year={2018}
}

@article{genevay2016stochastic,
  title={Stochastic optimization for large-scale optimal transport},
  author={Genevay, Aude and Cuturi, Marco and Peyr{\'e}, Gabriel and Bach, Francis},
  journal={Advances in Neural Information Processing Systems},
  volume={29},
  year={2016}
}

@inproceedings{korotinwasserstein,
  title={Wasserstein-2 generative networks},
  author={Korotin, Alexander and Egiazarian, Vage and Asadulaev, Arip and Safin, Alexander and Burnaev, Evgeny},
  booktitle={International Conference on Learning Representations},
  year = {2021}
}

@inproceedings{ijcai2019p305,
  title     = {Three-player {W}asserstein {GAN} via amortised duality},
  author    = {Dam, Nhan and Hoang, Quan and Le, Trung and Nguyen, Tu Dinh and Bui, Hung and Phung, Dinh},
  booktitle = {Proceedings of the Twenty-Eighth International Joint Conference on
               Artificial Intelligence, {IJCAI-19}},
  publisher = {International Joint Conferences on Artificial Intelligence Organization},
  pages     = {2202--2208},
  year      = {2019},
  month     = {7}
}

@article{leygonie2019adversarial,
  title={Adversarial computation of optimal transport maps},
  author={Leygonie, Jacob and She, Jennifer and Almahairi, Amjad and Rajeswar, Sai and Courville, Aaron},
  journal={arXiv preprint arXiv:1906.09691},
  year={2019}
}

@article{cao2019multi,
  title={Multi-marginal {W}asserstein {GAN}},
  author={Cao, Jiezhang and Mo, Langyuan and Zhang, Yifan and Jia, Kui and Shen, Chunhua and Tan, Mingkui},
  journal={Advances in Neural Information Processing Systems},
  volume={32},
  year={2019}
}

@inproceedings{liu2019wasserstein,
  title={Wasserstein {GAN} with quadratic transport cost},
  author={Liu, Huidong and Gu, Xianfeng and Samaras, Dimitris},
  booktitle={Proceedings of the IEEE/CVF international conference on computer vision},
  pages={4832--4841},
  year={2019}
}

@inproceedings{liu2018two,
  title={A two-step computation of the exact {GAN} {W}asserstein distance},
  author={Liu, Huidong and Xianfeng, Gu and Samaras, Dimitris},
  booktitle={International Conference on Machine Learning},
  pages={3159--3168},
  year={2018},
  organization={PMLR}
}

@inproceedings{wei2018improving,
  title={Improving the improved training of {W}asserstein {GAN}s: {A} Consistency term and its dual effect},
  author={Wei, Xiang and Gong, Boqing and Liu, Zixia and Lu, Wei and Wang, Liqiang},
  booktitle={International Conference on Learning Representations},
  year={2018}
}

@inproceedings{xie2019scalable,
  title={On scalable and efficient computation of large scale optimal transport},
  author={Xie, Yujia and Chen, Minshuo and Jiang, Haoming and Zhao, Tuo and Zha, Hongyuan},
  booktitle={International Conference on Machine Learning},
  pages={6882--6892},
  year={2019},
  organization={PMLR}
}

@article{korotin2021neural,
  title={Do neural optimal transport solvers work? a continuous {W}asserstein-2 benchmark},
  author={Korotin, Alexander and Li, Lingxiao and Genevay, Aude and Solomon, Justin M and Filippov, Alexander and Burnaev, Evgeny},
  journal={Advances in neural information processing systems},
  volume={34},
  pages={14593--14605},
  year={2021}
}

@inproceedings{rout2021generative,
  title={Generative modeling with optimal transport maps},
  author={Rout, Litu and Korotin, Alexander and Burnaev, Evgeny},
  booktitle={Proceedings of the International Conference on Learning Representations},
  year={2022}
}

@article{lu2020large,
  title={Large-scale optimal transport via adversarial training with cycle-consistency},
  author={Lu, Guansong and Zhou, Zhiming and Shen, Jian and Chen, Cheng and Zhang, Weinan and Yu, Yong},
  journal={arXiv preprint arXiv:2003.06635},
  year={2020}
}

@article{courty2016optimal,
  title={Optimal transport for domain adaptation},
  author={Courty, Nicolas and Flamary, R{\'e}mi and Tuia, Devis and Rakotomamonjy, Alain},
  journal={IEEE transactions on pattern analysis and machine intelligence},
  volume={39},
  number={9},
  pages={1853--1865},
  year={2016},
  publisher={IEEE}
}

@article{courty2017joint,
  title={Joint distribution optimal transportation for domain adaptation},
  author={Courty, Nicolas and Flamary, R{\'e}mi and Habrard, Amaury and Rakotomamonjy, Alain},
  journal={Advances in Neural Information Processing Systems},
  volume={30},
  year={2017}
}

@article{bernard2007optimal,
  title={Optimal mass transportation and Mather theory},
  author={Bernard, Patrick and Buffoni, Boris},
  journal={Journal of the European Mathematical Society},
  volume={9},
  number={1},
  pages={85--121},
  year={2007}
}

@book{Figalli2008OptimalTransport,
  author    = {Figalli, Alessio},
  title     = {Optimal Transportation and Action-Minimizing Measures},
  volume    = {8},
  publisher = {Edizioni della Normale},
  year      = {2008},
}

@book{villani2008optimal,
  title={Optimal transport: old and new},
  author={Villani, C{\'e}dric},
  volume={338},
  year={2008},
  publisher={Springer}
}

@book{villani2021topics,
  title={Topics in optimal transportation},
  author={Villani, C{\'e}dric},
  volume={58},
  year={2021},
  publisher={American Mathematical Soc.}
}
\bibliographystyle{amsplain}

\appendix
\section{Proof of consistency} \label{pf1}
We show that problem (\ref{learn_F}) is equivalent to problem (\ref{Problem_W}) by proving the following identity 
\begin{equation*}
    \inf_{F}\sup_{\phi: \text{Lip}} f_1(G)+ f_2(G,\phi) = 
    \inf_{G:W\left(G(1;\cdot)_{\#}\rho_a, \rho_b\right)=0}f_1(G),
\end{equation*}
where
\begin{align*}
    G(t;z) & = z + t F(z,t), \\
    f_1(G):& =
 \mathbb{E}\left[
     L\left(\frac{d G(U;Z)}{dt}\right)\right],\\
    f_2(G,\phi):& = \mathbb{E}\left[\phi\left(G(1;Z)\right)\right]-
    \mathbb{E}\left[\phi(Y)\right].
\end{align*}
Assume the min-max problem (\ref{learn_F}) admits a unique solution $(F^*,\phi^*)$, and define  
$$
G^*(t;z) := z+tF^*(z,t).
$$ 
Let $\phi$  be any Lipschitz function. By triangle inequality,    both
$\phi^*+\phi$ and $\phi^*-\phi$ are also Lipschitz functions. 
In particular, since $\phi^*+\phi$ is Lipschitz, we have
\begin{align*}
    & f_1(G^*)+ f_2(G^*,\phi^*+\phi)
    \le f_1(G^*)+ f_2(G^*,\phi^*)\\
    \Rightarrow & f_2(G^*,\phi^*+\phi) - f_2(G^*,\phi^*) \le 0. \\
    \Rightarrow & 
    \mathbb{E}\Bigr[\phi^*\left(G^*(1;Z)\right)+
    \phi\left(G^*(1;Z)\right)\Bigr]-
    \mathbb{E}\left[\phi^*(Y)+\phi(Y)\right]-
    \mathbb{E}\left[\phi^*\left(G^*(1;Z)\right)\right]+
    \mathbb{E}\left[\phi^*(Y)\right]\le 0\\
    \Rightarrow&\mathbb{E} \Bigr[  \phi\left(G^*(1;Z)\right)\Bigr]-
    \mathbb{E}\left[\phi(Y)\right]
    \le 0.
\end{align*}
Since $\phi^*-\phi$ is also a Lipschitz function, we can apply the same argument as above to obtain
\begin{equation*}
     \mathbb{E} \Bigr[ \phi\left(G^*(1;Z)\right)\Bigr]-
    \mathbb{E}\left[\phi(Y)\right]
    \ge 0,
\end{equation*}
which implies that 
\begin{equation*}
    f_2(G^*,\phi) = \mathbb{E} \Bigr[  \phi\left(G^*(1;Z)\right)\Bigr]-
    \mathbb{E}\left[\phi(Y)\right] = 0
     ~~ \text{for all Lipschitz } \phi.
\end{equation*}
By the Kantorovich-Rubinstein duality (\ref{KRduality}), we then have 
\begin{equation*}
    W\left(G^*(1;\cdot)_{\#}\rho_a, \rho_b\right) =  \sup_{\phi: \text{1-Lip}}
    \mathbb{E} \Bigr[  \phi\left(G^*(1;Z)\right)\Bigr]-
    \mathbb{E}\left[\phi(Y)\right] = \sup_{\phi: \text{1-Lip}}f_2(G^*,\phi) = 0
\end{equation*}
Finally, we conclude the proof by showing the equivalence of the min-max and constrained formulations:
\begin{align*}
     \inf_{F}\sup_{\phi: \text{Lip}} f_1(G)+ f_2(G,\phi)
    &=  \sup_{\phi: \text{Lip}}\inf_{F} f_1(G)+ f_2(G,\phi) \\
    &=   \sup_{\phi: \text{Lip}} f_1(G^*)+ f_2(G^*,\phi)\\
    &=  \sup_{\phi: \text{Lip}} f_1(G^*)\\
    &= f_1(G^*)\\
    &= \inf_{G:W\left(G(1;\cdot)_{\#}\rho_a, \rho_b\right)=0}f_1(G),
\end{align*}
where the third line follows from the assumption that $(F^*, \phi^*)$ is the unique saddle point and $f_2(G^*, \phi)=0$ for all Lipschitz 
$\phi$, as shown above.

\section{Details of numerical experiments}\label{detail_numerical}
In this section, we report the evolution of the objective functions over the training iteration in each experiment.
\subsection{Synthetic-1}
\begin{figure}
 \centering
 \includegraphics[width=0.8\textwidth]{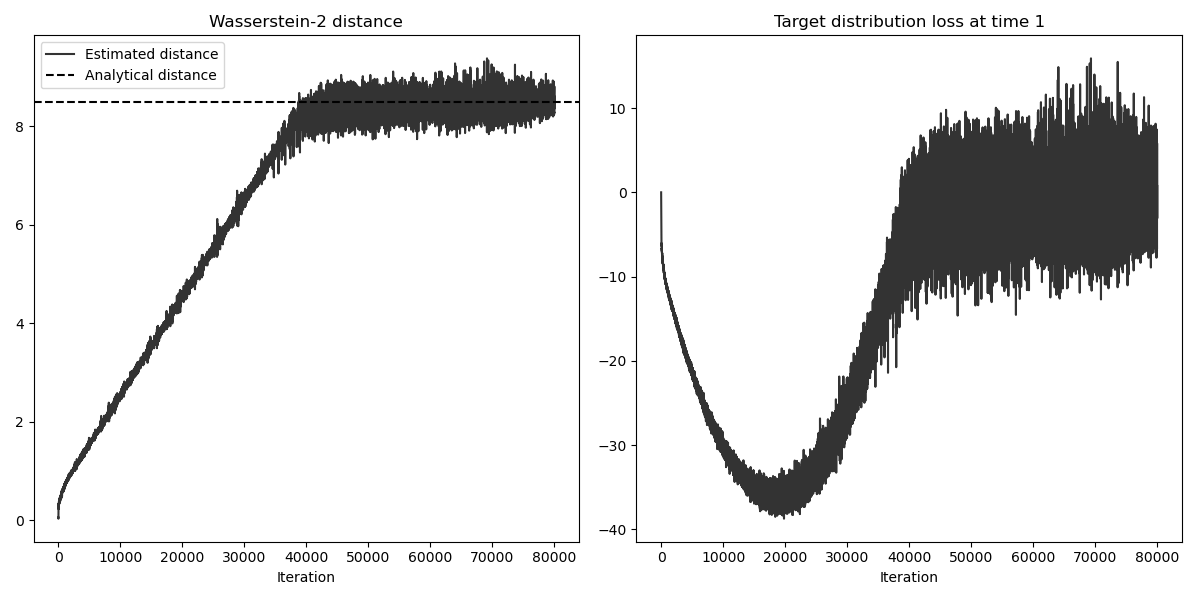}
 \caption{Synthetic-1, Phase 1: The left panel shows the estimated $W_2(\rho_a,
 \rho_b)$ over training  iterations; the dash line refers to the analytical value.
 The right panel shows the estimated $W_1(G_\theta(1;\cdot)_\#\rho_a,
 \rho_b)$ over iterations. }\label{Gau2D_loss}
\end{figure}

\begin{figure}
 \centering
 \includegraphics[width=0.8\textwidth]{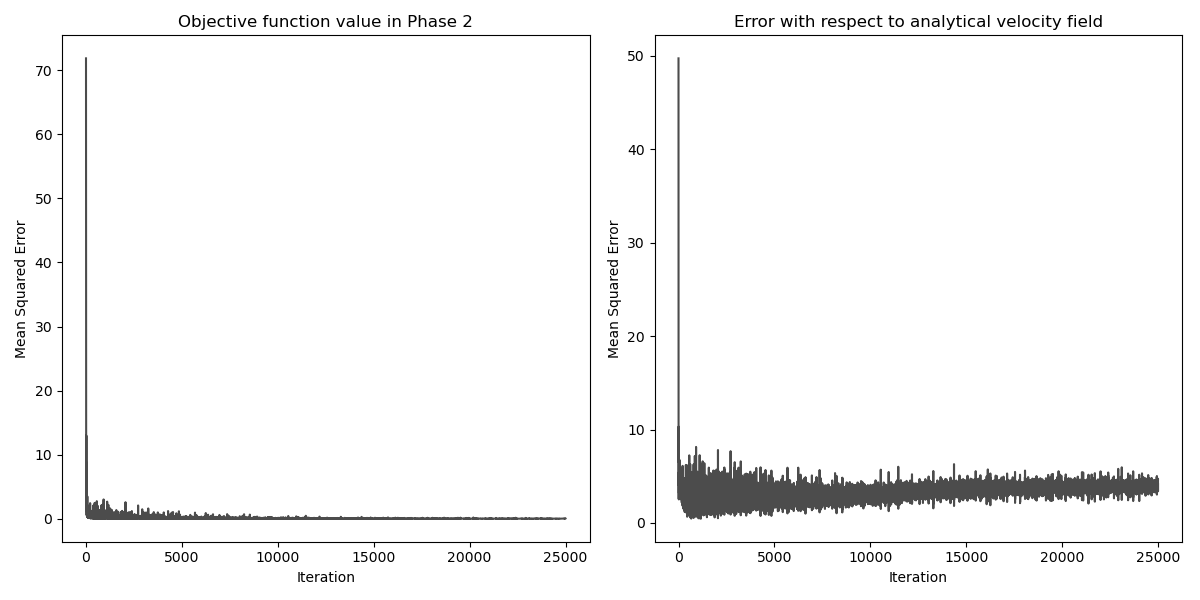}
 \caption{Synthetic-1, Phase 2: The left panel shows the loss of objective function value. 
 The right panel shows the mean squared error between the learned velocity field $v_\eta$ and the analytical one $v^*$.}\label{Gau2D_lossv}
\end{figure}

Figure \ref{Gau2D_loss} tracks individual terms of the objective function (\ref{learn_Ftheta}) in Phase 1 over the training iterations.
Fig \ref{Gau2D_lossv} presents the loss of the objective function (\ref{learn_v}) in Phase 2 over the training iterations. 
Since the analytical velocity field $v^*$ is available for this case, so we additionally report the mean squared error between $v_\eta(x,t)$ and $v^*(x,t)$ over the domain $x\in(-3,10.9)\times(-3,10.9)$ and $t\in(0,1)$. 
\subsection{Synthetic-2}
\begin{figure}
 \centering
 \includegraphics[width=0.8\textwidth]{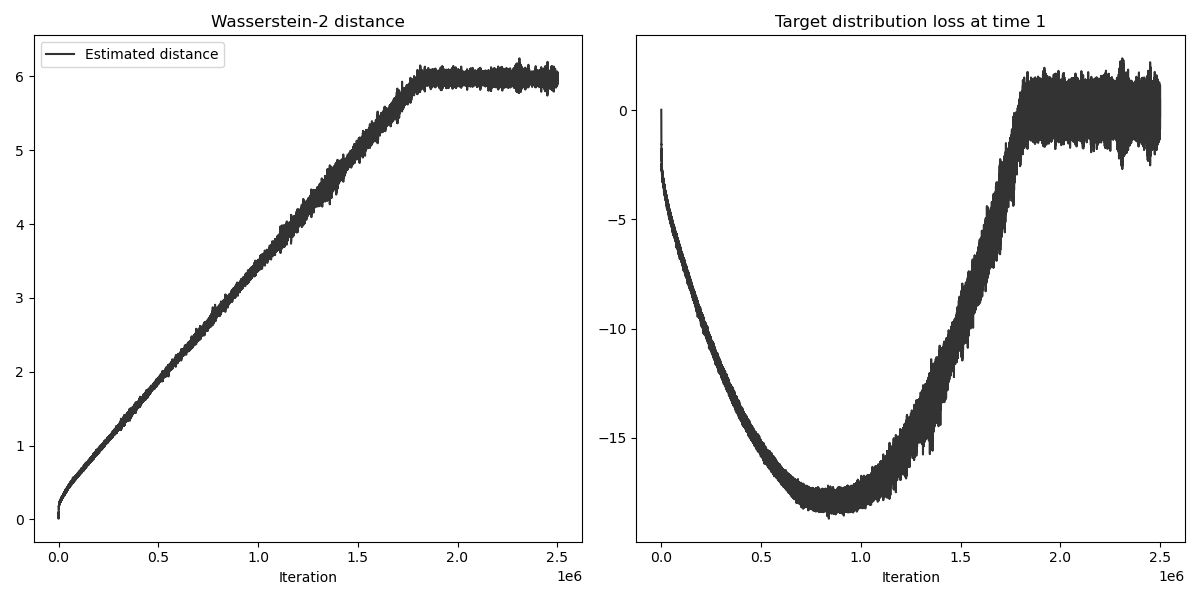}
 \caption{Synthetic-2, Phase 1: The left panel shows the estimated $W_2(\rho_a,
 \rho_b)$ over training iterations; the dash line refers to the analytical value.
 The right panel shows the estimated $W_1(G_\theta(1;\cdot)_\#\rho_a,
 \rho_b)$ over iterations.}\label{MixG2D_loss}
\end{figure}

\begin{figure}
 \centering
 \includegraphics[width=0.4\textwidth]{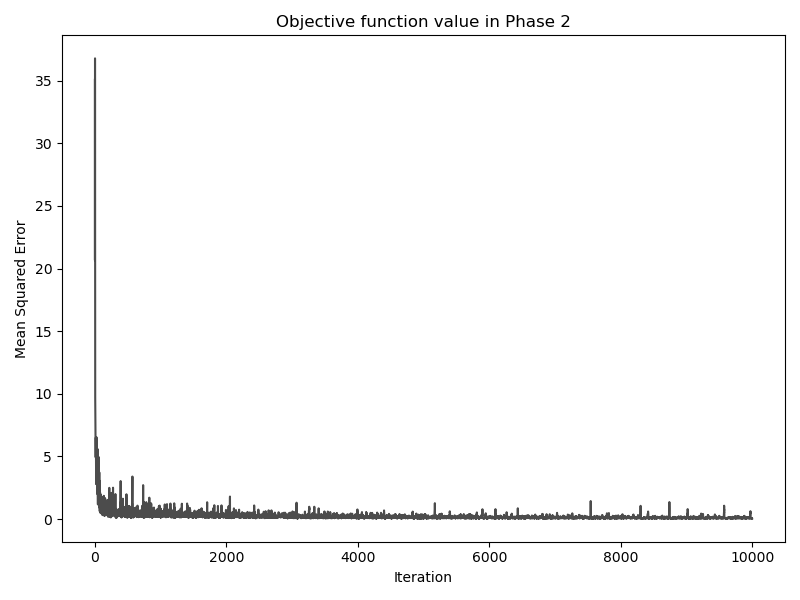}
 \caption{Synthetic-2, Phase 2: Loss of the objective function value. 
 }\label{MixedG2D_lossv}
\end{figure}

Fig \ref{MixG2D_loss} tracks individual terms of the objective function (\ref{learn_Ftheta}) in Phase 1 over the training iterations.
Fig \ref{MixedG2D_lossv} presents the loss of the objective function (\ref{learn_v}) in Phase 2 over training iterations. 
\subsection{Synthetic-3}
\begin{figure}
 \centering
 \includegraphics[width=0.8\textwidth]{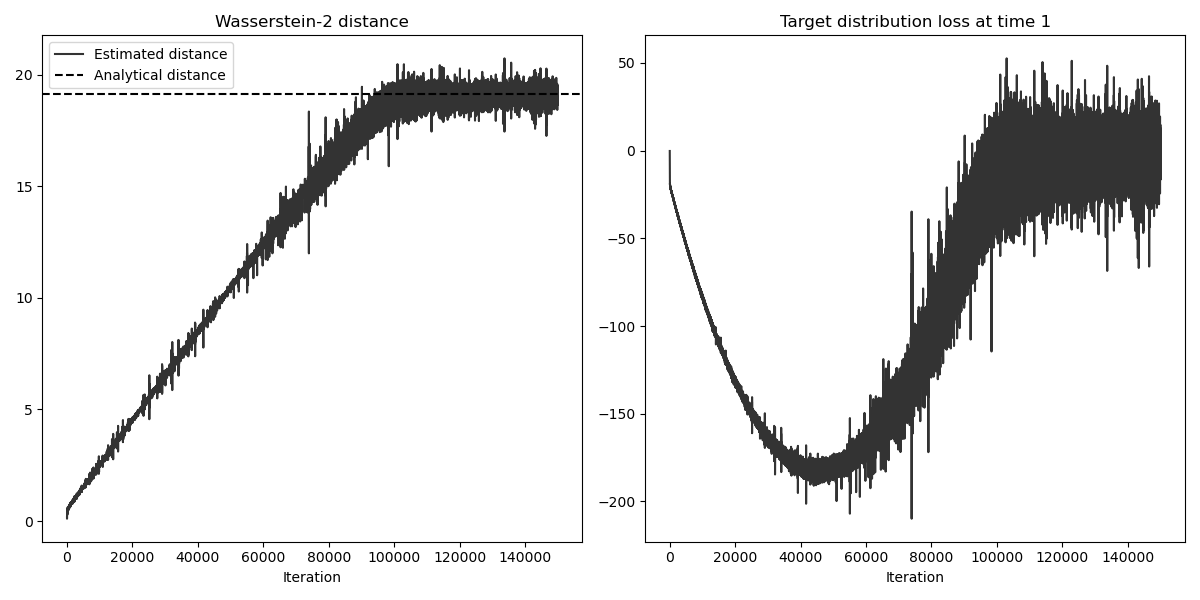}
 \caption{Synthetic-3, Phase 1: The left panel shows the estimated $W_2(\rho_a,
 \rho_b)$ over training  iterations; the dash line refers to the analytical value.
 The right panel shows the estimated $W_1(G_\theta(1;\cdot)_\#\rho_a,
 \rho_b)$ over iterations.}\label{Gau10D_loss}
\end{figure}

\begin{figure}
 \centering
 \includegraphics[width=0.8\textwidth]{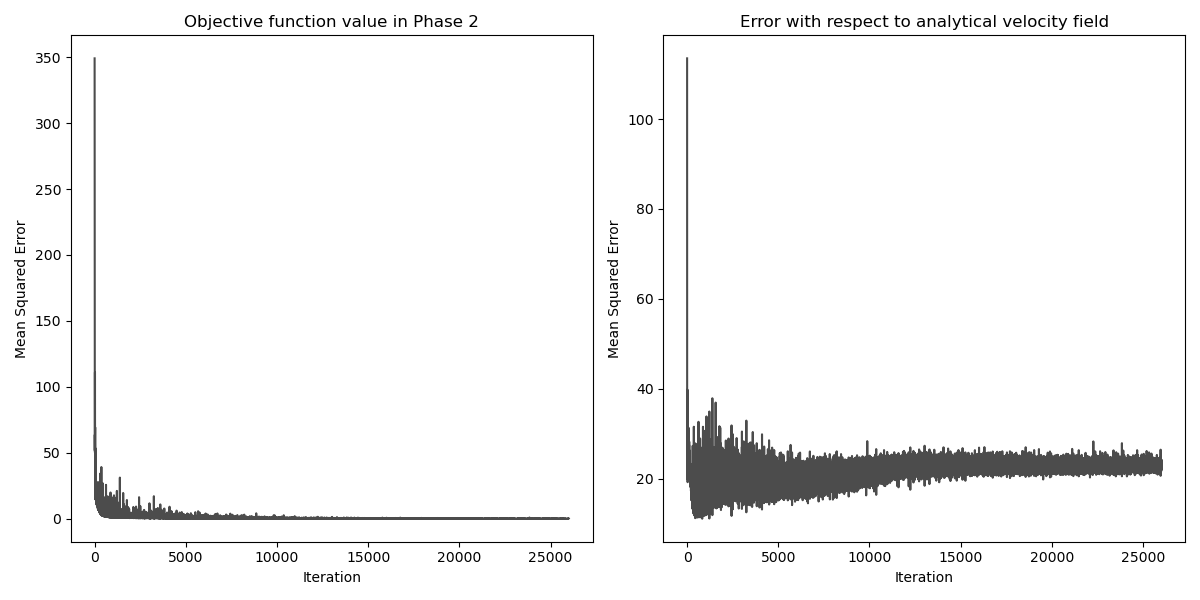}
 \caption{Synthetic-3, Phase 2: The left panel shows the loss of objective function value. 
 The right panel shows the mean squared error between the learned velocity field $v_\eta$ and the analytical one $v^*$ }\label{Gau10D_lossv}
\end{figure}

Fig \ref{Gau10D_loss} tracks individual terms of the objective function (\ref{learn_Ftheta}) in Phase 1 over the training iterations.
Fig \ref{Gau10D_lossv} presents the loss of the objective function (\ref{learn_v}) in Phase 2 over the training iterations. As in Synthetic-1, the analytical velocity field $v^*$ is available for this case, so we additionally report the mean squared error between $v_\eta(x,t)$ and $v^*(x,t)$ over 100000 random points in a 10-dimensional hyperrectangle containing the means of the two functions. 

\subsection{Synthetic-4}
\begin{figure}
 \centering
 \includegraphics[width=0.8\textwidth]{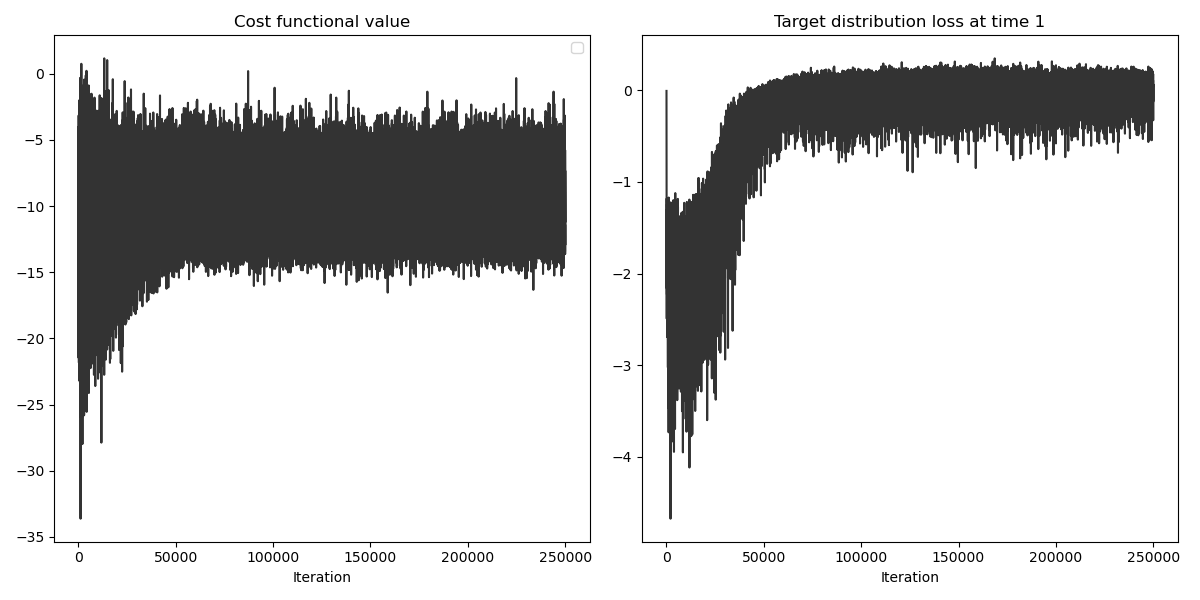}
 \caption{Synthetic-4: The left panel shows the estimated cost over training  iterations.
 The right panel shows the estimated $W_1(G_\theta(1;\cdot)_\#\rho_a,
 \rho_b)$ over iterations.}\label{HO2D_loss}
\end{figure}

\begin{figure}
 \centering
 \includegraphics[width=0.4\textwidth]{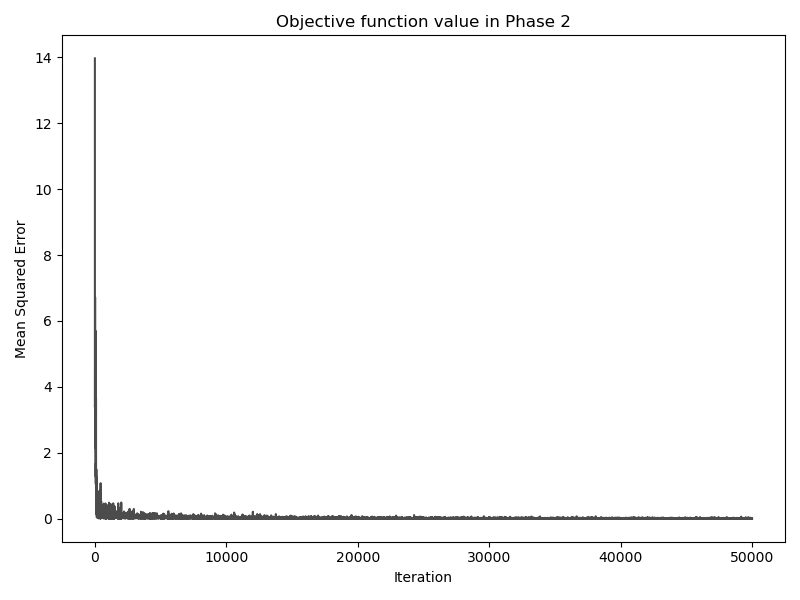}
 \caption{Synthetic-4: Loss of the objective function value. }\label{HO2D_lossv}
\end{figure}

Fig \ref{HO2D_loss}  tracks the objective function loss (\ref{learn_Ftheta}) in Phase 1 over training iterations.  
Fig \ref{HO2D_lossv} presents the corresponding loss of the objective function (\ref{learn_v}) in Phase 2 during training iterations. 
\subsection{Real data (MNIST)}
\begin{figure}
 \centering
 \includegraphics[width=0.8\textwidth]{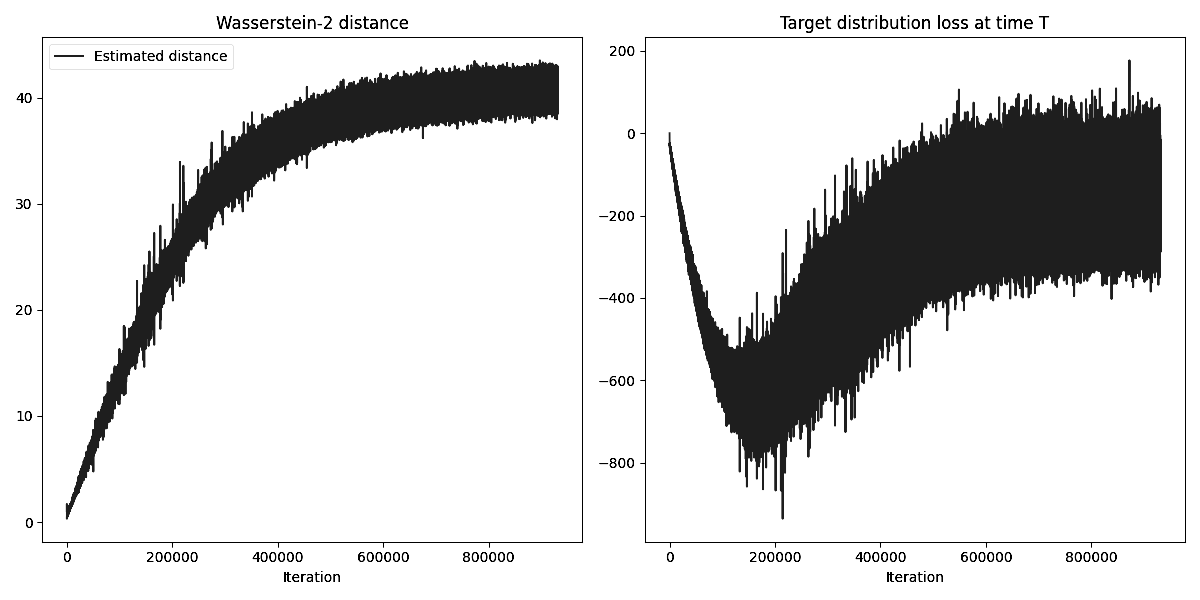}
 \caption{Realistic case: The left panel shows the estimated $W_2(\rho_a,
 \rho_b)$ over training iterations. The right panel shows the estimated $W_1(G_\theta(1;\cdot)_\#\rho_a,
 \rho_b)$ over iterations.}\label{MNIST_loss}
\end{figure}

\begin{figure}
\centering
\includegraphics[width=0.4\textwidth]{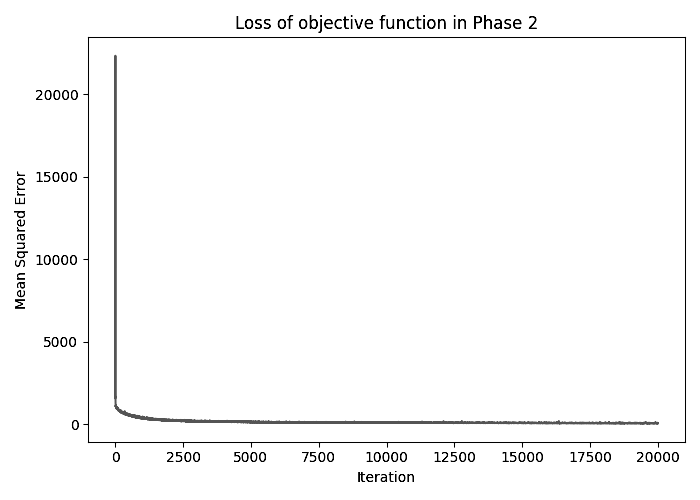}
\caption{Realistic case, Phase 2: Loss of the objective function value.}\label{MNIST_lossv}
\end{figure}

Fig \ref{MNIST_loss}  tracks the loss of the objective function (\ref{learn_Ftheta}) in Phase 1 over the training iterations.
Fig \ref{MNIST_lossv} tracks the loss of the objective function (\ref{learn_v}) in Phase 2 over iterations. 

\end{document}